\documentclass[10pt,twocolumn,letterpaper]{article}

\usepackage[pagenumbers]{cvpr} 

\usepackage{multirow} 

\newcommand{\abbr}{\text{VIAFormer}\xspace}

\usepackage{mathtools} 
\usepackage{algorithm}
\usepackage{algpseudocode}
\usepackage{amsmath}
\usepackage{amssymb}
\usepackage{booktabs}
\usepackage{threeparttable}

\definecolor{cvprblue}{rgb}{0.21,0.49,0.74}
\usepackage[pagebackref,breaklinks,colorlinks,allcolors=cvprblue]{hyperref}

\def\paperID{} 
\def\confName{CVPR}
\def\confYear{2026}

\title{VIAFormer: Voxel-Image Alignment Transformer for\\ High-Fidelity Voxel Refinement}

\author{
  Tiancheng Fang$^{1}$ \quad
  Bowen Pan$^{2}$ \quad
  Lingxi Chen$^{1}$ \quad
  Jiangjing Lyu$^{2}$ \\
  Chengfei Lyu$^{2}$ \quad
  Chaoyue Niu$^{1}$\thanks{Corresponding author.} \quad
  Fan Wu$^{1}$
   \\
  $^{1}$Shanghai Jiao Tong University \quad
  $^{2}$Alibaba Group \\
  {\tt\small \{fangtiancheng,sjtu\_chenlingxi,Rvince,fwu\}@sjtu.edu.cn} \\
  {\tt\small \{bowen.pbw,jiangjing.ljj,chengfei.lcf\}@alibaba-inc.com}
}

\begin{document}
\maketitle
\begin{abstract}
We propose VIAFormer, a \textbf{V}oxel-\textbf{I}mage \textbf{A}lignment Trans-\textbf{former} model designed for Multi-view Conditioned Voxel Refinement—the task of repairing incomplete noisy voxels using calibrated multi-view images as guidance. Its effectiveness stems from a synergistic design: an Image Index that provides explicit 3D spatial grounding for 2D image tokens, a Correctional Flow objective that learns a direct voxel-refinement trajectory, and a Hybrid Stream Transformer that enables robust cross-modal fusion. Experiments show that VIAFormer establishes a new state of the art in correcting both severe synthetic corruptions and realistic artifacts on the voxel shape obtained from powerful Vision Foundation Models. Beyond benchmarking, we demonstrate VIAFormer as a practical and reliable bridge in real-world 3D creation pipelines, paving the way for voxel-based methods to thrive in large-model, big-data wave.
\end{abstract}
\section{Introduction}
\label{sec:intro}

Voxel grids are a fundamental representation in 3D processing pipelines. As inputs, they are crucial for high-fidelity generative models that produce complex shapes \cite{trellis25}. As outputs, voxel grids are commonly generated by various reconstruction methods, such as Vision Foundation Models (VFMs) \cite{Dense3R25, vggt25, pi3_25, MapAnything25} and 3D scanning. However, a significant quality gap exists between these two use cases. For example, while VFMs can easily produce coarse voxel grids with surface noise and missing undersides, these imperfect grids are not directly suitable as inputs for high-demand generative models. Bridging this gap requires a dedicated process, which we refer to as ``Voxel Refinement''. 


Along voxel refinement, existing methods such as DiffComplete \cite{DiffComplete23}, PatchComplete \cite{PatchComplete22}, and WSSC \cite{wssc25} suffer from fundamental design limitations that restrict their effectiveness. A key drawback is their exclusive reliance on geometry-based approaches, which lack the benefit of valuable multi-modal supervision, such as from multi-view images. Additionally, their underlying model architectures are often based on convolutional networks operating in explicit 3D space, suffering from poor scalability, where resolutions scaling from a low $32^3$ to a moderate $64^3$ lead to prohibitive memory costs. This architectural choice is often coupled with a restrictive data paradigm: models are typically trained on small, category-specific datasets and require object labels, hindering their ability to generalize to large-scale, diverse, and unlabeled data. 

\begin{figure}[t]
  \centering   \includegraphics[width=1.0\linewidth]{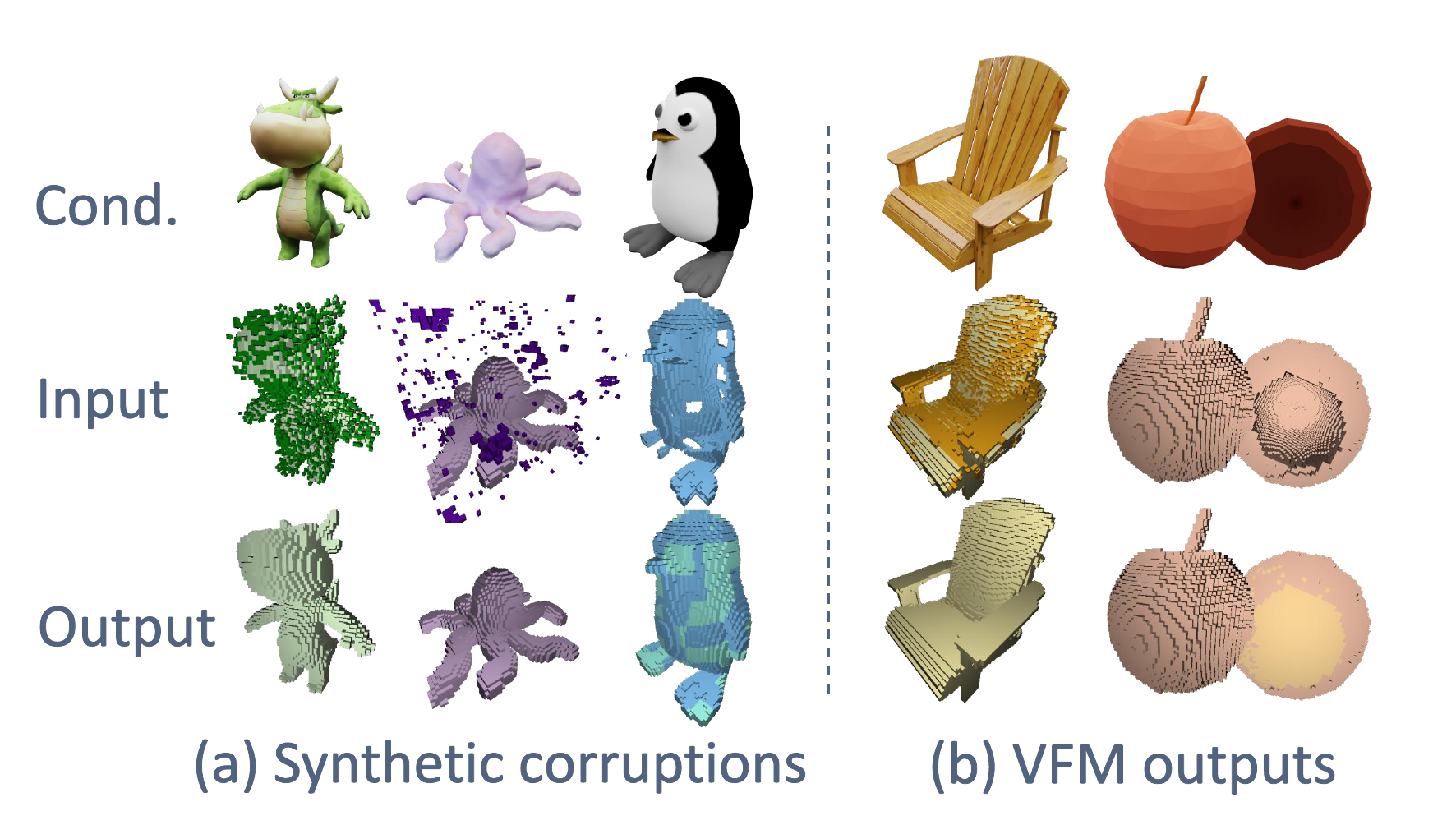}

   \caption{\textbf{The Conditioned Voxel Refinement Task.} \abbr learns to refine a wide range of imperfections guided by multi-view images (Cond.). It handles both (a) various synthetic corruptions and (b) realistic degradations from Vision Foundation Models' outputs, producing clean and complete voxel grids.}
   \label{fig:task_intro}
\end{figure}

To address these shortcomings, we propose a new task: \textbf{Conditioned Voxel Refinement}, where the goal is to repair an incomplete noisy voxel under the guidance of multi-view images. We formalize this task for a large-scale context, focusing on mid-resolution ($64^3$) grids and category-agnostic datasets. As illustrated in Fig.~\ref{fig:task_intro}, the objective is to develop a versatile model capable of handling a broad range of imperfections. These include synthetic corruptions, such as surface noise, floating artifacts, and structural holes (Fig.~\ref{fig:task_intro}a), as well as realistic degradations from VFMs' outputs, like large occluded regions and uneven surfaces (Fig.~\ref{fig:task_intro}b).

In this work, we introduce {\bf\abbr}, a Voxel-Image Alignment Transformer. It leverages a correctional flow objective to learn a direct pathway from noisy latent to its clean counterpart rather than generating from scratch. To ensure precise multimodal guidance, we introduce the Image Index, which establishes explicit 3D spatial grounding for 2D image tokens. These components are integrated within a Hybrid Stream Transformer that effectively fuses the geometric and visual cues, enabling not only to denoise and complete local geometry but also to infer and restore global structures based on multi-view evidence.


The contributions of this work are threefold:
\begin{itemize}
    \item We formalize the task of Conditioned Voxel Refinement in a modern, large-scale, and multi-modal setting, addressing a key gap in 3D pipelines.
    \item We propose \abbr, featuring a Correctional Flow objective, an Image Index, and a Hybrid Stream Transformer to enable robust, image-guided voxel refinement.
    \item We demonstrate through extensive experiments that \abbr achieves state-of-the-art performance on both VFM-derived and synthetic noise benchmarks.
\end{itemize}

\section{Releated Work}
\label{sec:related_works}

\subsection{Voxel Completion and Refinement}
\label{sec:2_1}
The use of 3D voxel grids, prized for their structural simplicity, is prevalent in 3D shape generation. Early approaches to voxel completion focused on recovering missing geometry from partial observations by leveraging learned shape priors. These include methods based on local patch statistics (e.g., PatchComplete~\cite{PatchComplete22}) and those utilizing global shape codes derived from autoencoders (e.g., AutoSDF~\cite{autosdf2022}). More recently, a dominant trend has emerged toward 3D-native diffusion models (e.g., DiffComplete~\cite{DiffComplete23}, SDFusion~\cite{SDFusion23}), which achieve high-fidelity generation from random noise. However, these methods often suffer from limited generalization to unseen categories due to their reliance on training-specific shape distributions.
A recent approach, WSSC~\cite{wssc25}, tackles this limitation by exploiting a pre-collected prior bank to guide voxel-based completion. While WSSC demonstrates improved generalization to new categories, its performance is still constrained by the completeness and diversity of the available priors, as well as the absence of image-based conditioning signals.
In contrast, our method harnesses the rich semantic and geometric priors encoded in a VFM as an initial prior, and further integrates image-conditioned features to effectively bridge partial inputs with complete structures.

\subsection{Image to 3D Generative Models}
\label{sec:2_2}

Since diffusion models surpassed GANs~\cite{NEURIPS2021_49ad23d1}, a large body of 3D generation research has adopted denoising diffusion~\cite{pmlr-v37-sohl-dickstein15,ho2020denoising,song2020denoising} or advanced flow matching~\cite{flowmatching23,flowmatchingcode24}. Some work with the idea of lifting 2D to 3D ~\cite{dreamfusion23,lin2023magic3d,wang2023prolificdreamer,shi2023MVDream,long2023wonder3d} can exploit 2D prior, but the quality of generated 3D assets is generally not high.
Another line of work based on 3D native generative model ~\cite{3DShape2VecSet,trellis25} directly learn the conditional generation process from a large-scale 3D dataset, showing strong performance under various conditioning signals. We cast voxel refinement as a 3D generation task and adopt the 3D-native generation framework: given multi-view images and an initial incomplete voxel grid, then output a refined 3D voxel shape.

\subsection{Conditional Control in 3D Generative Models}
\label{sec:2_3}
In practice, generative models are typically conditioned on user-specified inputs to produce desired outputs, making conditional control a critical factor in determining performance.
For U-Net~\cite{ronneberger2015u} based architecture such as DiffComplete~\cite{DiffComplete23}, multi-level conditional information is injected through an auxiliary network (e.g., ControlNet~\cite{zhang2023adding}).
Alternatively, Transformer~\cite{vaswani2017attention} based approaches~\cite{3DShape2VecSet,zhang2024clay,one2345pp24,MeshFormer24,trellis25,Hi3DGen25,reconviagen25,wssc25,hunyuan3d22025tencent,lai2025hunyuan3d25highfidelity3d,li2025step1x,roblox2025cube,lu2025orientation} typically stack multiple transformer blocks equipped with cross-attention to integrate conditioning signals.
While both ControlNet and cross-attention enable decent conditional generation, the granularity and fidelity of control remain suboptimal, primarily because the correspondence between generated 3D tokens and conditional 2D tokens is not explicitly learned or well aligned.
Our method enhances the semantic and geometric alignment between 3D voxel and 2D image tokens through a unified coordinate and space via prior camera parameters. Such an explicit structural guidance strengthens token-level interaction across modalities, improving conditional controllability and generation performance.

\begin{figure*}[t] 
    \centering \includegraphics[width=\textwidth]{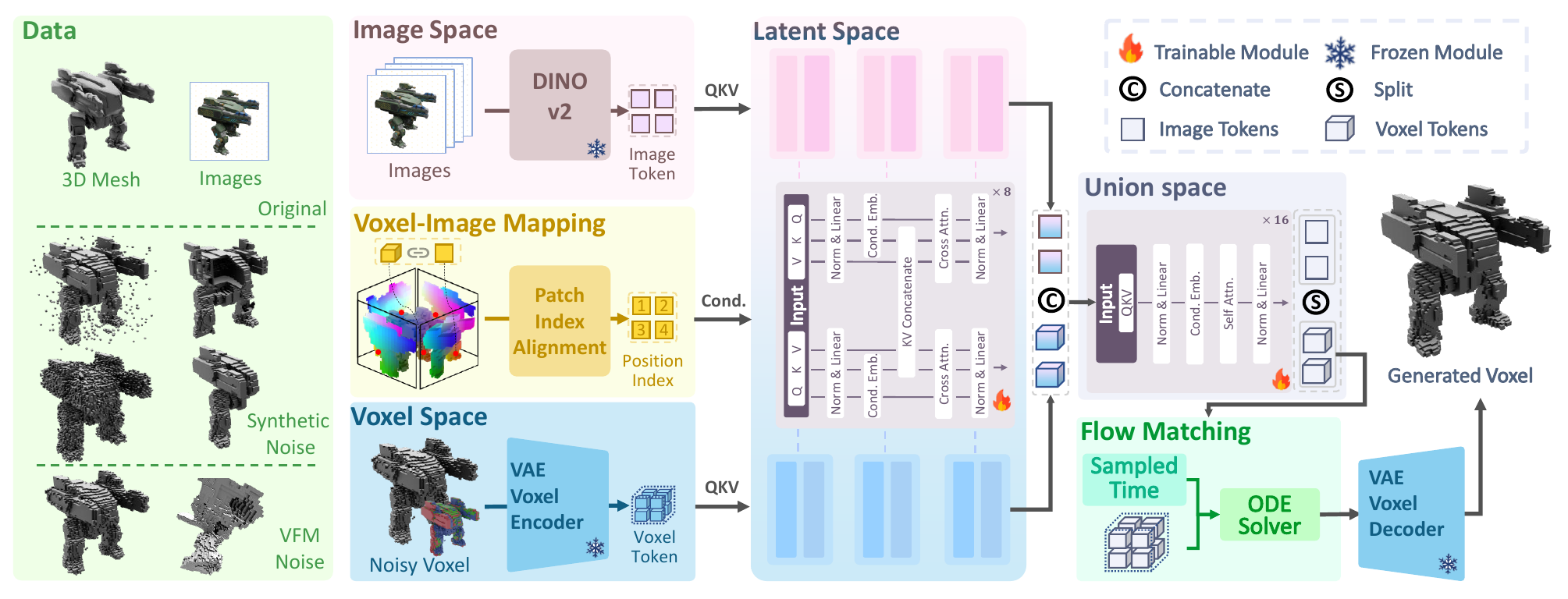} 
    \caption{\textbf{Overview of \abbr Architecture.} An incomplete noisy voxel grid (derived from VFMs or synthetic corruptions) and multi-view images are encoded into latent tokens. Our Image Index provides explicit 3D spatial locations to the 2D image patch tokens, creating a Union-Space for effective cross-modal fusion within a Hybrid Stream Transformer. The transformer, trained with a Correctional Flow objective, predicts a refinement velocity vector. This vector guides an ODE solver \cite{flowmatchingcode24} to produce a clean latent, which the VAE decoder converts into the final refined voxel grid.}
    \label{fig:pipeline}
\end{figure*}

\section{Task Formulation}
\label{sec:problem_formulation}


We focus on Conditioned Voxel Refinement: refining an incomplete noisy voxel $\tilde{v}$ into a high-quality voxel $\hat{v}$ using guidance from multi-view images $\{I_i\}_{i=1}^S$. We formulate the refinement process as learning a model $F_\theta$ where:
\begin{equation}
\label{eq:problem_formulation}
\hat{v} = F_\theta \left( \tilde{v}, \{I_i\}_{i=1}^S, c \right).
\end{equation}
The initial grid $\tilde{v}$ serves as a powerful but flawed geometric prior, representing a typical output from many 3D workflows. Examples include incomplete grids from LiDAR scans, noisy voxelizations of geometry estimated by VFMs~\cite{vggt25, pi3_25, MapAnything25}, or even native 3D assets that are inherently incomplete, such as those missing an underside.
The term $c$ represents any additional, helpful conditioning information. In this work, we demonstrate that estimated camera poses $\{ \mathbf{\tilde{T}}_i \}_{i=1}^S$ can serve as $c$ to facilitate the crucial alignment between 3D voxel tokens and 2D image tokens. The overall goal is to design an effective architecture $F_\theta$ that can learn to interpret the global structure from $\tilde{v}$ while leveraging detailed visual evidence from the images to correct geometric errors and complete missing regions. 


\section{Method}\label{sec:method}
The overview of the proposed VIAFormer is illustrated in Fig.~\ref{fig:pipeline}. The pipeline begins by encoding the primary inputs into a latent token space. The initial noisy grid $\tilde{v}$ is passed through a pretrained sparse structure VAE encoder $\mathcal{E}_V$ \cite{trellis25}, while the set of guiding images $\{I_i\}$ are processed by a powerful DINOv2 encoder $\mathcal{E}_I$ \cite{dinov2_24} to obtain patch-based features:
\begin{equation}
\label{eq:encoding}
z_V = \mathcal{E}_V(\tilde{v}); \quad \{z_{I,i}\}_{i=1}^S = \{\mathcal{E}_I(I_i)\}_{i=1}^S
\end{equation}
where $z_V$ is the geometric latent and $\{z_{I,i}\}$ are the image patch tokens.

At the heart of our method is a Hybrid Stream Transformer that learns to reverse the geometric degradation. It follows a correctional trajectory defined by a Flow Matching objective (Sec.~\ref{sec:correctional_flow}), learning to map the noisy latent $z_V$ back to its clean counterpart. This refinement process is made possible by our proposed Image Index (Sec.~\ref{sec:patch_index}), a novel mechanism that establishes direct spatial grounding for 2D image tokens. By explicitly bridging this modal gap, the Image Index allows our transformer (Sec.~\ref{sec:hybrid_stream_transformer}) to effectively fuse the two information streams within a unified feature space. The process concludes when the pretrained VAE decoder~\cite{trellis25} transforms the refined latent code back into the final voxel grid $\hat{v}$.

\subsection{Correctional Flow}
\label{sec:correctional_flow}
Probabilistic generative models, such as those based on Diffusion or Flow Matching (FM) \cite{flowmatching23}, typically learn to transform samples from a simple Gaussian distribution into a target data distribution. In our case, however, we are given a coarse grid $\tilde{v}$ that, while noisy, contains rich structural information about the target geometry. The most intuitive approach, therefore, is not to start from random noise, but to learn a direct path from the noisy latent representation $z_V = \mathcal{E}_V(\tilde{v})$ to its clean counterpart $z_{\text{gt}} = \mathcal{E}_V(v_{\text{gt}})$. We term this specialized formulation a \textbf{Correctional Flow}.

We implement this by defining a linear path between the two latents, $z_t = (1-t)z_V + t \cdot z_{\text{gt}}$ for $t \in [0,1]$. The network $f_\theta$ is then trained to predict the constant velocity field of this path, i.e., the ``correction vector'' $(z_{\text{gt}} - z_V)$. This leads to the following objective:
\begin{equation}
\label{eq:fm_loss}
\mathcal{L}_{\text{FM}} = \mathbb{E}_{t, \tilde{v}, v_{\text{gt}}, c} \left[ \left\| f_\theta(z_t, t, c) - (z_{\text{gt}} - z_V) \right\|_2^2 \right].
\end{equation}
Here, the conditioning set $c$ comprises all guiding information provided to the transformer. This includes the image tokens $\{z_{I,i}\}$ which are spatially grounded via our Image Index (Sec.~\ref{sec:patch_index}). This formulation directly tasks the model with geometric refinement, a more constrained problem than generation from scratch.

\subsection{The Necessity of a Voxel-Image Union-Space}
\label{sec:necessity_of_union_space}
\begin{figure}[]
  \centering   \includegraphics[width=1.0\linewidth]{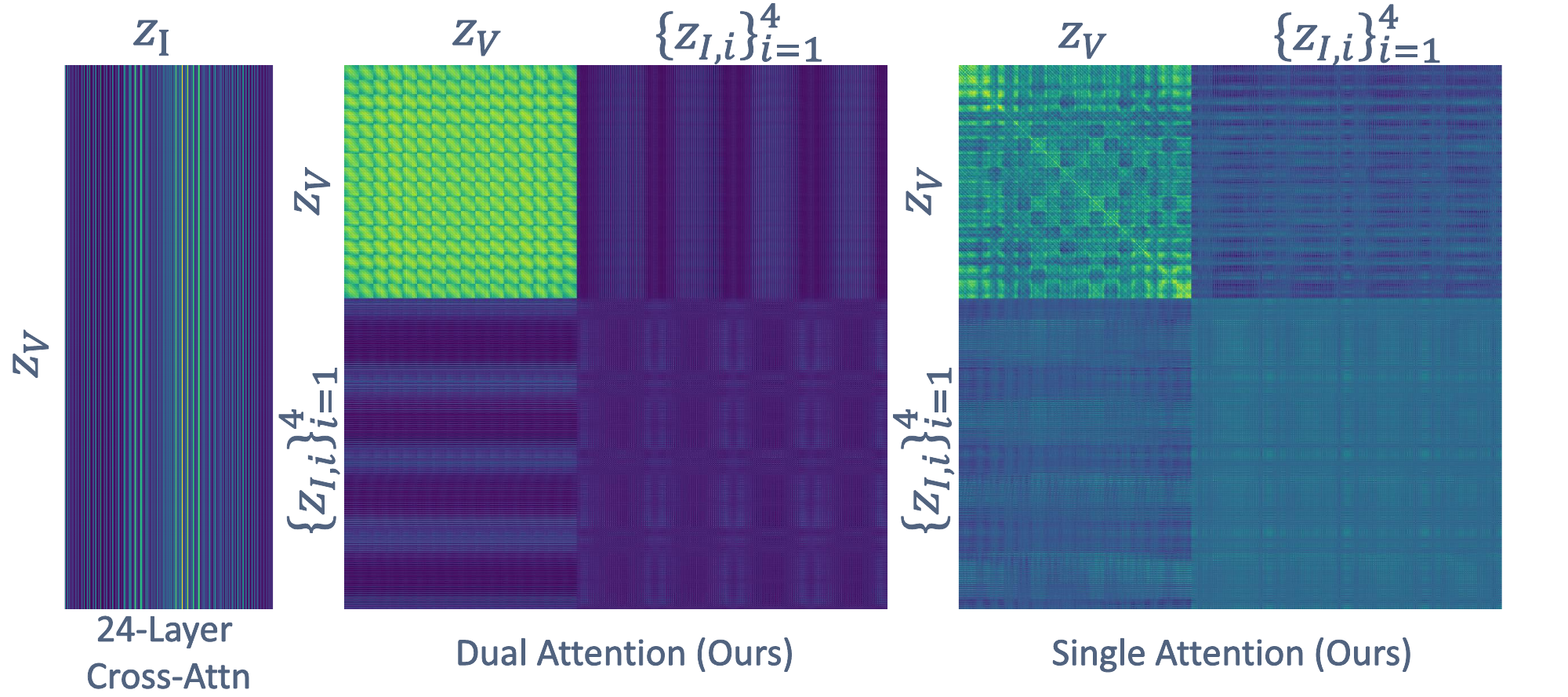}

   \caption{\textbf{Attention Map Visualization.} (Left) Standard cross-attention exhibits Attention Collapse, failing to learn spatial correspondence. (Middle and Right) The Dual Attention and Single Attention of our model, respectively, operate within the Voxel-Image Union-Space to enable structured, bidirectional attention.}
   \label{fig:attention_map}
\end{figure}

Cross-attention, a common mechanism for injecting conditioning, proves ineffective for our task. As confirmed in Table~\ref{tab:main_results}, cross-attention offers negligible improvement over a geometry-only self-attention version. 

To diagnose the intrinsic nature, we visualize the attention map, which illustrates the influence weights ($\text{softmax}(QK^T/\sqrt{d_k})$) assigned by each voxel token (query) to every image token (key). As shown in Fig.~\ref{fig:attention_map}, the resulting map exhibits a pattern of largely uniform stripes, indicating that the influence of any given image token is spread almost equally across all voxel tokens—a phenomenon we term \textbf{Attention Collapse}. To confirm that Attention Collapse is a fundamental issue in cross-modal interaction, rather than one of weak conditioning or a simplistic interface, we replaced the image condition with powerful, pre-trained VGGT \cite{vggt25} features and tested two adapters for their integration: a 3-layer MLP and a separate 8-layer self-attention module. As detailed in Table~\ref{tab:main_results}, neither approach produced meaningful improvement. These failures demonstrate that simply enhancing the conditioning signal or adapter complexity is insufficient. The core problem is structural: standard cross-attention lacks a shared spatial basis, causing the model to disregard positional information and treat the image context as a single, global feature.

Therefore, we propose Voxel-Image Union-Space, a new paradigm that establishes this shared basis by assigning a 3D coordinate to every 2D and 3D token. These coordinates are then converted into sinusoidal embeddings \cite{vaswani2017attention} and injected into their respective tokens via RoPE \cite{RoPE21,FLUX25}. This ensures that physical proximity in 3D space translates directly into embedding similarity, effectively bridging the gap between modalities. This structural alignment biases the attention mechanism: the attention scores between spatially proximate tokens are naturally higher, thus preventing attention collapse and enabling a meaningful, dynamic, and bidirectional exchange of information.

\subsection{Union-Space Token Alignment}
\label{sec:patch_index}
\begin{figure}[t]
  \centering   \includegraphics[width=1.01\linewidth]{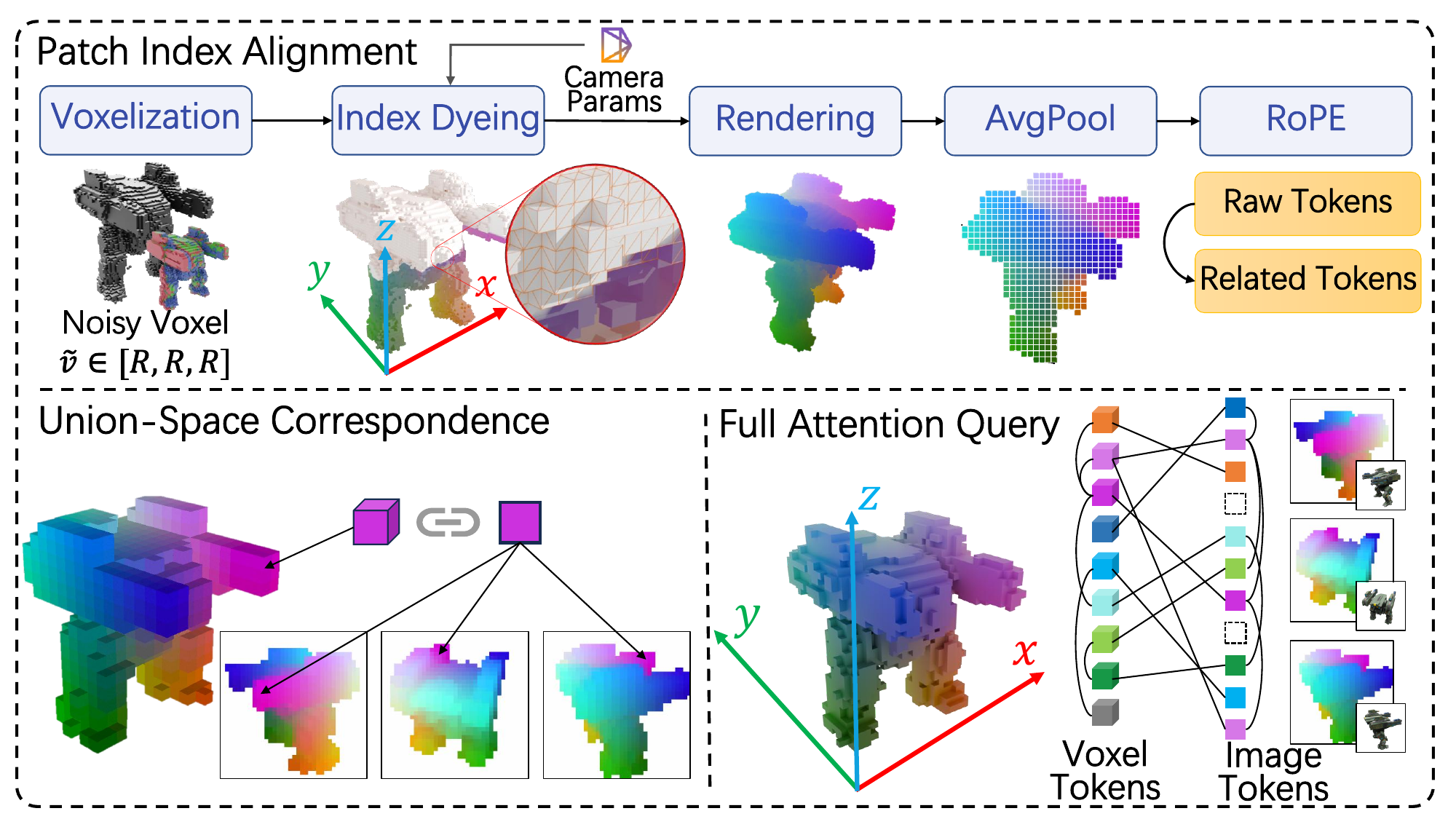}

  \caption{\textbf{Illustration of the Image Index and Union-Space Correspondence.} The Image Index is generated by rendering the input voxel grid $\tilde{v}$ with its 3D coordinates encoded as colors, thereby assigning a spatial location to each 2D image patch token (top). This explicit geometric grounding enables spatially-aware cross-attention between voxel and image tokens, allowing the model to perform effective multi-view guided refinement (bottom).}
   \label{fig:img_index}
\end{figure}
To establish the Voxel-Image Union-Space, the primary challenge is enabling effective cross-stream attention between the voxel and image modalities. We propose to solve this by actively aligning the two token streams. Since each voxel token naturally possesses 3D coordinates, we achieve alignment by assigning a corresponding 3D coordinate to each image token. This process directly embeds 3D spatial information onto the 2D tokens, which endows them with explicit positional awareness and creates a “lookup key” for queries from the voxel stream.

We realize this lookup key as the \textbf{Image Index}, a set of 3D coordinates generated for each image token via a efficient rendering-based process (Fig.~\ref{fig:img_index}). First, we generate a mesh via a simple triangulation of $\tilde{v}$ and encode the 3D integer coordinate of each source voxel directly into the face colors. Next, this coordinate-encoded mesh is rendered from each camera pose $\{\mathbf{\tilde T}_i\}$ to produce 2D index maps, where background pixels are marked with a null value. Finally, we partition these maps into patches (with the same patch size as DINOv2 \cite{dinov2_24}) and compute a single floating-point 3D coordinate for each image token by average pooling the non-null pixel coordinates within its patch. The resulting sequence of coordinates forms the Image Index, providing the shared geometric grounding for all subsequent cross-modal attention.

Since the Image Index is generated from the noisy input grid $\tilde{v}$ and camera poses $\{\mathbf{\tilde T}_i\}$, it inherits their geometric inaccuracies, establishing what is fundamentally a coarse spatial prior. However, we argue that even this coarse alignment is vastly superior to none at all. It provides the essential “spatial handshake” needed to bootstrap the attention mechanism, enabling the transformer to locate relevant image features and begin the refinement process.

\subsection{Hybrid Stream Transformer}
\label{sec:hybrid_stream_transformer}

We implement the Union-Space concept with a Hybrid Stream Transformer, whose architecture follows the design of OmniControl \cite{OminiControl24}. The model consists of 24 transformer blocks organized into two stages: an initial dual-stream stage for bidirectional alignment, followed by a single-stream stage for global fusion.

The initial 8 blocks are dual-stream blocks, designed to establish the shared geometric grounding. Within each block, the voxel latent $z_V$ and the multi-view latents $\{z_{I,i}\}$ are first projected into queries ($Q$), keys ($K$), and values ($V$) using non-weight-sharing MLPs to preserve their distinct features. We then construct a unified key-value space by concatenating the key and value tensors from both streams:
\begin{gather}
    K_{\text{union}} = \text{Concat}(K_V, K_{I,1}, \cdots, K_{I,S}), \\
    V_{\text{union}} = \text{Concat}(V_V, V_{I,1}, \cdots, V_{I,S}).
\end{gather}
This enables true bidirectional attention, where both the voxel and image streams query this shared space, allowing them to co-evolve:
\begin{gather}
    z_V \coloneqq z_V + \text{Attention}(Q_V, K_{\text{union}}, V_{\text{union}}), \\
    z_{I,i} \coloneqq z_{I,i} + \text{Attention}(Q_{I,i}, K_{\text{union}}, V_{\text{union}}).
\end{gather}

Following this initial alignment stage, the voxel and image streams, now possessing a shared geometric grounding, are concatenated into a single unified sequence:
\begin{equation}
z_{\text{unified}} = \text{Concat}(z_V, z_{I,1}, \cdots, z_{I,S}).
\end{equation}
This unified sequence is then processed by the final 16 single-stream blocks. These standard transformer blocks apply self-attention uniformly across the combined set of tokens, facilitating global feature fusion and allowing the model to reason about the scene as a holistic entity.

\subsection{Training Data Synthesis}
\label{sec:training_synthesis}

We combine two complementary data sources. The first is generated by processing multi-view images through Pi3 \cite{pi3_25} and voxelizing the resulting point cloud. The second, a procedural noise pipeline, generates three main, combinable types of corruptions: surface noise, which applies random perturbations near the object's boundary; volumetric artifacts, which add both isolated floaters and clustered noise in space far from the boundary; and coarse-level masking, which randomly discards entire voxel blocks. In addition to these, we apply a more drastic half-space removal to create severe structural occlusions, rigorously testing the model's reliance on visual information.

Under the dual-source strategy, by training on a 1:1 mixture of sources, we prevent the model from learning spurious correlations from the VFM data, such as the tendency to hallucinate a ``phantom base'' on objects that should be open from below, a common artifact of datasets dominated by top-down views. Instead, the procedural corruptions force the model to develop a more robust and generalizable geometric understanding.

\section{Experiments}
\label{sec:experiments}

\begin{table*}[t]
\centering
\caption{Comprehensive quantitative comparison for refining VFM-derived artifacts on the Toys4k \cite{Toys4K21} and Dora \cite{Dora25} datasets. We report Mean and Standard Deviation for all metrics. $\uparrow$ indicates higher is better, and $\downarrow$ indicates lower is better.}
\label{tab:main_results}
\resizebox{0.95\textwidth}{!}{%
\begin{tabular}{l cc|cc cc|cc cc}
\toprule
\multirow{2}{*}{\textbf{Method}} & \multicolumn{2}{c|}{\multirow{2}{*}{\textbf{Category}}} & \multicolumn{4}{c|}{\textbf{Toys4k} \cite{Toys4K21}} & \multicolumn{4}{c}{\textbf{Dora} \cite{Dora25}} \\
\cmidrule(lr){4-7} \cmidrule(lr){8-11}
& \multicolumn{2}{c|}{} & \multicolumn{2}{c}{\textbf{IoU} $\uparrow$} & \multicolumn{2}{c}{\textbf{Chamfer Dist.} $\downarrow$} & \multicolumn{2}{c}{\textbf{IoU} $\uparrow$} & \multicolumn{2}{c}{\textbf{Chamfer Dist.} $\downarrow$} \\
\cmidrule(lr){2-3} \cmidrule(lr){4-5} \cmidrule(lr){6-7} \cmidrule(lr){8-9} \cmidrule(lr){10-11}
& \textbf{Views} & \textbf{Init} & \textbf{Mean} & \textbf{Std} & \textbf{Mean} & \textbf{Std} & \textbf{Mean} & \textbf{Std} & \textbf{Mean} & \textbf{Std} \\
\midrule
Pi3 Voxelization \cite{pi3_25} & 4 & -- & 0.3959 & 0.0897 & 0.0193 & 0.0130 & 0.4244 & 0.1129 & 0.0192 & 0.0109 \\
\midrule
DiffComplete \cite{DiffComplete23} & 0 & Pi3 & 0.4012 & 0.0974 & 0.0194 & 0.0131 & 0.3842 & 0.1024 & 0.0218 & 0.0114 \\
PatchComplete \cite{PatchComplete22} & 0 & Pi3 & 0.4249 & 0.1397 & 0.0191 & 0.1367 & 0.3637 & 0.0163 & 0.0232 & 0.0211 \\
WSSC \cite{wssc25} & 0 & Pi3 & 0.3580 & 0.1554 & 0.0321 & 0.0481 & 0.3319 & 0.1749 & 0.0303 & 0.0420 \\
SDFusion \cite{SDFusion23} & 1 & Pi3 & 0.4071 & 0.1048 & 0.0197 & 0.0135 & 0.4070 & 0.1182 & 0.0210 & 0.0120 \\
Trellis \cite{trellis25} & 4 & Noise & 0.3332 & 0.1998 & 0.0164 & 0.0131 & 0.2269 & 0.2267 & 0.0895 & 0.0841  \\
\midrule
24-Layer Self-Attn & 0 & Pi3 & 0.4111 & 0.0968 & 0.0184 & 0.0131 & 0.4285 & 0.0117 & 0.0190 & 0.0111 \\
24-Layer Cross-Attn & 1 & Pi3 & 0.4255 & 0.1015 & 0.0175 & 0.0131 & 0.4356 & 0.1195 & 0.0182 & 0.0120 \\
24-Layer Cross-Attn (VGGT-cond. + 3MLP) & 4 & Pi3 & 0.4055 & 0.0942 & 0.0188 & 0.0130 & 0.4243 & 0.1073 & 0.0191 & 0.0108 \\
24-Layer Cross-Attn (VGGT-cond. + 8Self-Attn) & 4 & Pi3 & 0.4076 & 0.0951 & 0.0188 & 0.0131 & 0.4236 & 0.1033 & 0.0193 & 0.0111 \\
\midrule
\abbr (w/o Image Index) & 4 & Pi3 & 0.4176 & 0.0973 & 0.0180 & 0.0131 & 0.4418 & 0.1276 & 0.0181 & 0.0110 \\
\abbr (w/o Correctional Flow) & 4 & Noise & 0.3102 & 0.0995 & 0.0259 & 0.0161 & 0.2675 & 0.1203 & 0.0326 & 0.0164 \\
\textbf{\abbr (Ours)} & \textbf{4} & \textbf{Pi3} & \textbf{0.4460} & 0.0988 & \textbf{0.0163} & 0.0126 & \textbf{0.4585} & 0.1273 & \textbf{0.0172} & 0.0104 \\
\bottomrule
\end{tabular}
}
\end{table*}

\subsection{Experimental Setup}
\label{sec:experimental_setup}

\textbf{Data.} Our training data comprises approximately 478k 3D assets from ObjaverseXL \cite{ObjaverseXL23}, ABO \cite{ABO22}, 3D-FUTURE \cite{3DFuture21}, and HSSD \cite{HSSD24}. Evaluation is performed on the held-out Toys4k \cite{Toys4K21} and Dora dataset \cite{Dora25} to ensure fair comparison.

\noindent\textbf{Pre-processing.} Following the protocol of Trellis \cite{trellis25}, we render 150 views for each asset, primarily using RGB images but switching to normal maps for assets with problematic materials. From these views, we select $S=4$ images as input $\{I_i\}_{i=1}^S$ via a K-Means clustering strategy on the cameras' angular positions to ensure diversity. The ground-truth grid $v_{gt}$ is obtained by normalizing and voxelizing the asset's mesh in the resolution $R=64$.

\noindent\textbf{Implementation Details.}
The model size of the proposed \abbr is 0.61B and is trained with AdamW \cite{loshchilov2017decoupled} optimizer ($\text{lr}=3\times10^{-4}$) on 16 H20 GPUs over 7 days. 

\noindent\textbf{Evaluation Metrics.} To assess performance, we measure volumetric accuracy with Intersection over Union (IoU) and surface fidelity with Chamfer Distance.

\subsection{Baselines}
We comprehensively benchmark \abbr against a range of methods, categorized as follows:
\begin{itemize}
\item \textbf{VFM Baseline:} The point cloud predicted by the pre-trained Pi3 model \cite{pi3_25} is directly converted to voxel grid. This baseline represents the performance of our noisy input and serves as a reference.
\item \textbf{Geometry-Only Methods:} These models operate solely on the voxel input without image guidance. We consider classic shape completion methods DiffComplete \cite{DiffComplete23}, PatchComplete \cite{PatchComplete22} and WSSC \cite{wssc25}.
\item \textbf{Image-Conditioned Methods:} These models use a single image as guidance. We compare against SDFusion \cite{SDFusion23} and a cross-attention variant of Trellis \cite{trellis25}.
\end{itemize}

\noindent\textbf{Fair Comparison via Baseline Adaptation.}
A direct, out-of-the-box comparison with many baselines is infeasible and uninformative \cite{holopart25}. These methods were originally designed for different tasks (e.g., completing clean, partial shapes) or representations (e.g., operating in explicit $32^3$ voxel space). To isolate the impact of core architectural designs rather than differences in data representation or training objectives, we adapt all baselines to the correctional generation setting. The key modifications are:
\begin{itemize}
\item \textbf{Latent Space Operation:} We integrate all methods with same pre-trained VAE \cite{trellis25}, enabling them to operate on $64^3$ resolution within a compressed latent space.
\item \textbf{Re-trained:} All models are re-trained to predict the velocity field for our correctional flow objective, ensuring a consistent training paradigm.
\item \textbf{Bug Fixes \& Compatibility:} We corrected minor bugs present in the official code \cite{PatchComplete22,wssc25} and made necessary compatibility adjustments for our training framework. 
\end{itemize}
This rigorous adaptation protocol creates stronger, more relevant baselines and ensures that our performance gains are attributable to \abbr's superior architecture for multi-view guided refinement.

\subsection{Main Results}\label{sec:main_results}

\begin{figure*}[t] 
    \centering 
    \includegraphics[width=0.92\textwidth]{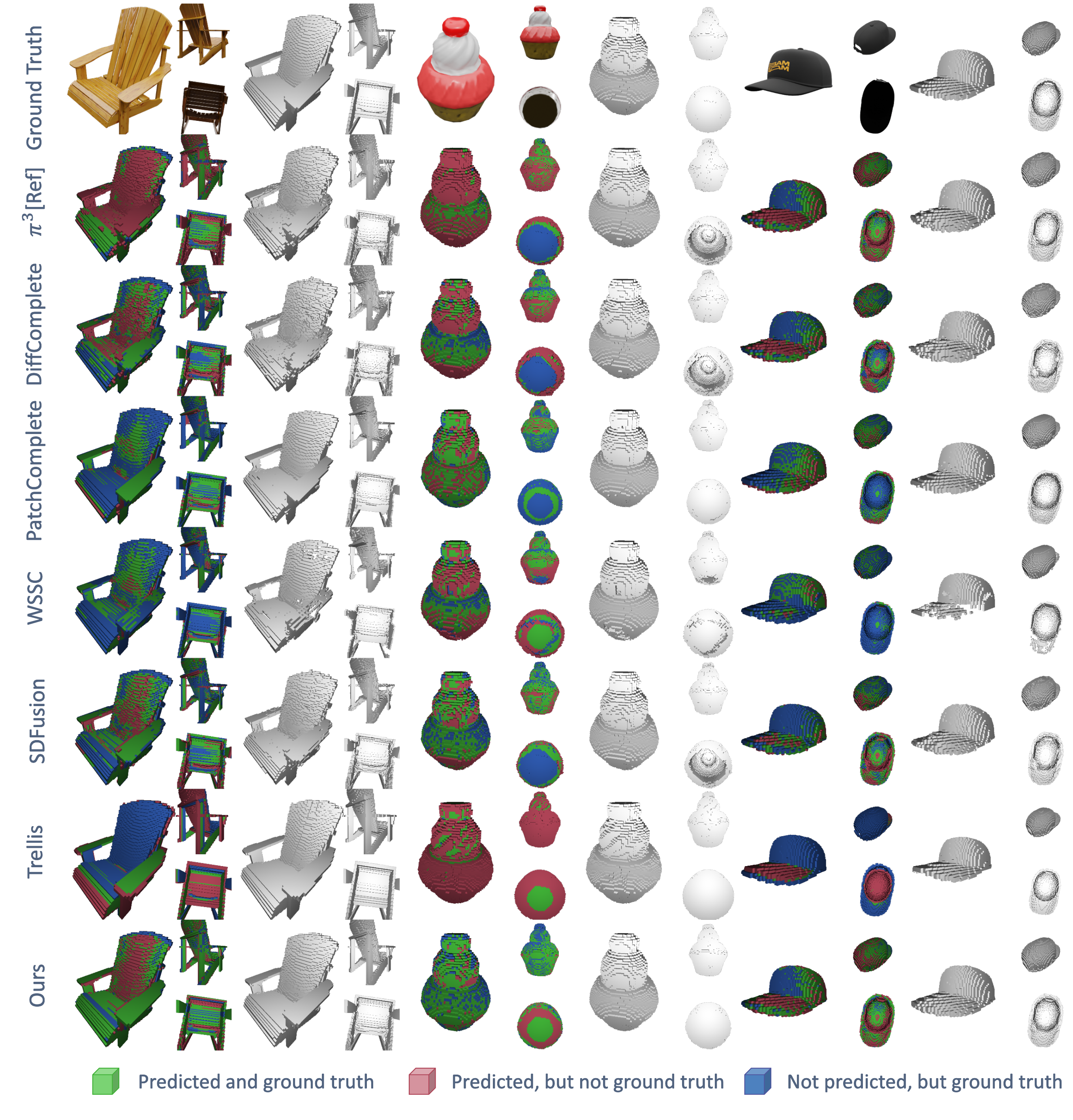} 
    \caption{\textbf{Visual Comparison of Voxel Refinement on VFM-Derived Inputs.} The error map highlights true positives (green), false positives (red), and false negatives (blue). Compared to the input (Pi3 \cite{pi3_25}) and other methods, \abbr consistently generates more complete and cleaner geometry with substantially fewer artifacts.}
    \label{fig:main_results}
\end{figure*}

\begin{table}[t]
\centering
\caption{Focused quantitative comparison on the synthetically corrupted Toys4k \cite{Toys4K21} dataset with baselines. We report Mean and Standard Deviation for all metrics.}
\label{tab:focused_results_template}
\resizebox{\columnwidth}{!}{%

\begin{tabular}{lc|cc cc}
\toprule
& & \multicolumn{4}{c}{\textbf{Toys4k} \cite{Toys4K21}} \\
\cmidrule(lr){3-6}
\textbf{Method} & \textbf{Condition} & \multicolumn{2}{c}{\textbf{IoU} $\uparrow$} & \multicolumn{2}{c}{\textbf{Chamfer Dist.} $\downarrow$} \\
\cmidrule(lr){3-4} \cmidrule(lr){5-6}
& & \textbf{Mean} & \textbf{Std} & \textbf{Mean} & \textbf{Std} \\
\midrule
Input Reference & -- & 0.4666 & 0.1224 & 0.0315 & 0.0227 \\
\midrule
DiffComplete \cite{DiffComplete23} & no & 0.2127 & 0.1188 & 0.1079 & 0.0698 \\
PatchComplete \cite{PatchComplete22} & no & 0.0414 & 0.0706 & 0.2117 & 0.1222 \\
WSSC \cite{wssc25} & no & 0.0036 & 0.0159 & 0.6389 & 0.2061 \\
SDFusion \cite{SDFusion23} & 1 view & 0.2620 & 0.1281 & 0.0980 & 0.0748 \\
24-Layer Self-Attn & no & 0.2776 & 0.1521 & 0.0868 & 0.0671 \\
24-Layer Cross-Attn & 1 view & 0.2590 & 0.1347 & 0.0766 & 0.0793 \\
\textbf{\abbr (Ours)} & 4 views & \textbf{0.8580} & 0.0912 & \textbf{0.0027} & 0.0032 \\
\bottomrule
\end{tabular}%
}
\end{table}

We present both quantitative and qualitative results to demonstrate the effectiveness of \abbr, which achieves new state-of-the-art performance. On the task of correcting VFM-derived artifacts (Table~\ref{tab:main_results}), our method improves IoU by 5.0\% on Toys4k and 3.4\% on Dora, with corresponding reductions in Chamfer Distance. The gains are even more striking on synthetically corrupted data (Table~\ref{tab:focused_results_template}), where \abbr boosts IoU by a remarkable 39.1\% on the Toys4k dataset.

\begin{figure}[]
  \centering   \includegraphics[width=0.95\linewidth]{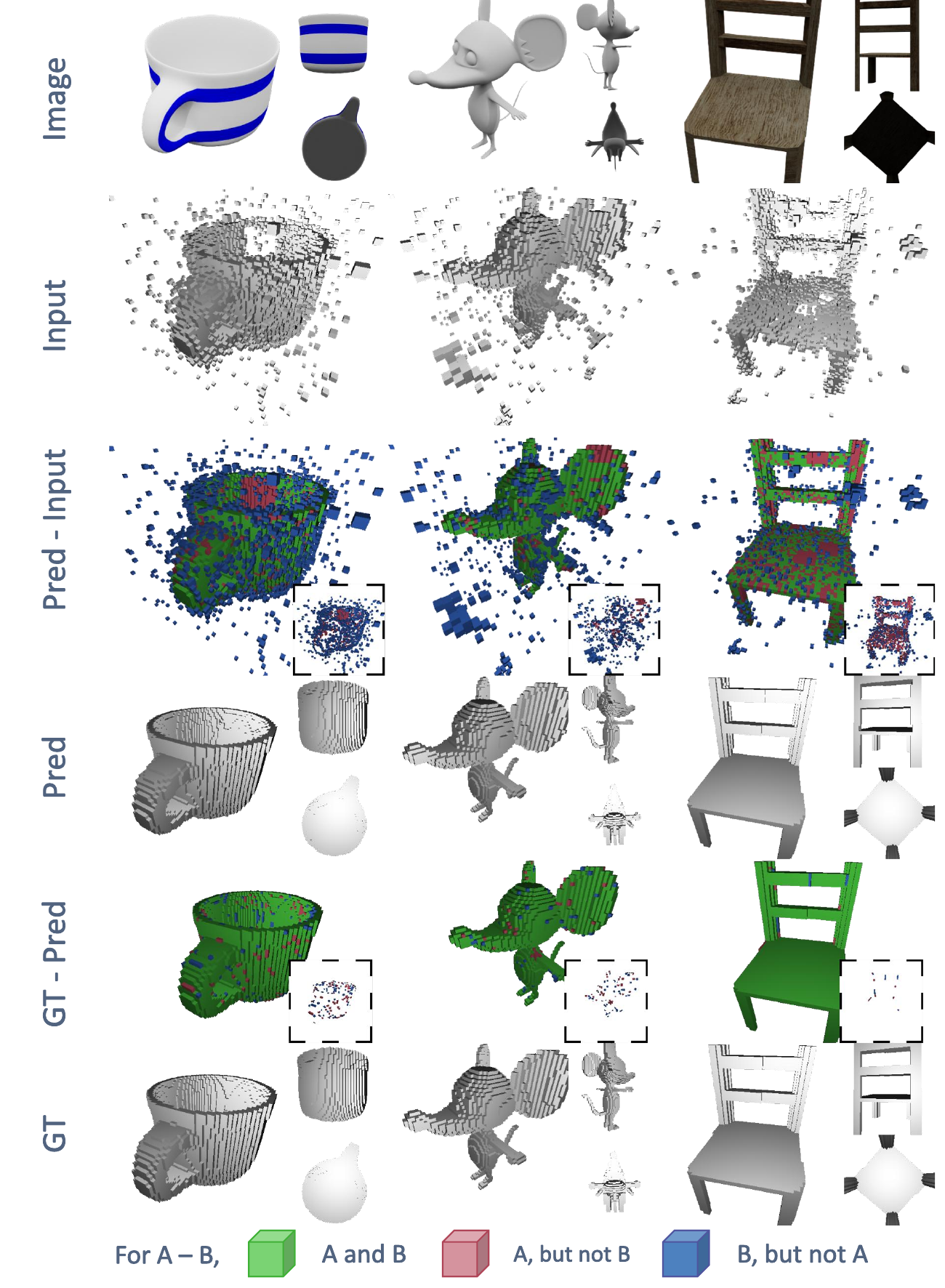}

   \caption{\textbf{Refinement Results on Synthetic Noise.} We break down how \abbr corrects inputs with severe combined corruptions. The ``Pred - Input'' row visualizes the model's corrections, while the ``GT - Pred'' row, along with the inset error-only views, reveals the residual errors of the final prediction against the ground truth (GT).}
   \label{fig:noised_results}
\end{figure}

The numerical superiority is visually corroborated in Fig. \ref{fig:main_results} and \ref{fig:noised_results}. Our method showcases powerful geometric refinement capabilities, successfully eliminating fine-grained surface noise, removing spurious floaters, and completing large missing regions like object undersides.

\subsection{Ablation Study}
\label{sec:ablation_study}
Our ablation studies, presented in Table~\ref{tab:main_results} on VFM-derived artifacts, validate the synergistic contributions of \abbr's core components. We first demonstrate the inadequacy of standard cross-attention: a series of strong baselines, including single-image and VGGT-conditioned variants, all offer negligible improvement over a geometry-only model. This failure confirms our prior analysis that an explicit spatial grounding mechanism, a Voxel-Image Union-Space, is necessary.

We then validate our two key components. First, replacing the Image Index with a standard ViT-style positional encoding (\abbr (w/o Image Index)) leads to a clear performance drop (e.g., IoU from 0.446 to 0.418 on Toys4k), proving that our explicit spatial grounding is what enables meaningful multi-modal fusion. Second, we validate our Correctional Flow formulation. Training from pure noise (\abbr (w/o Correctional Flow)) instead of refining the VFM-initialized latent causes performance to plummet, with IoU dropping from 0.446 to 0.310. Collectively, these results affirm that each component of \abbr is essential for achieving state-of-the-art refinement.

\subsection{Impact of the Number of Input Views}
\label{sec:further_analysis}

\begin{figure}[t]
  \centering   \includegraphics[width=0.92\linewidth]{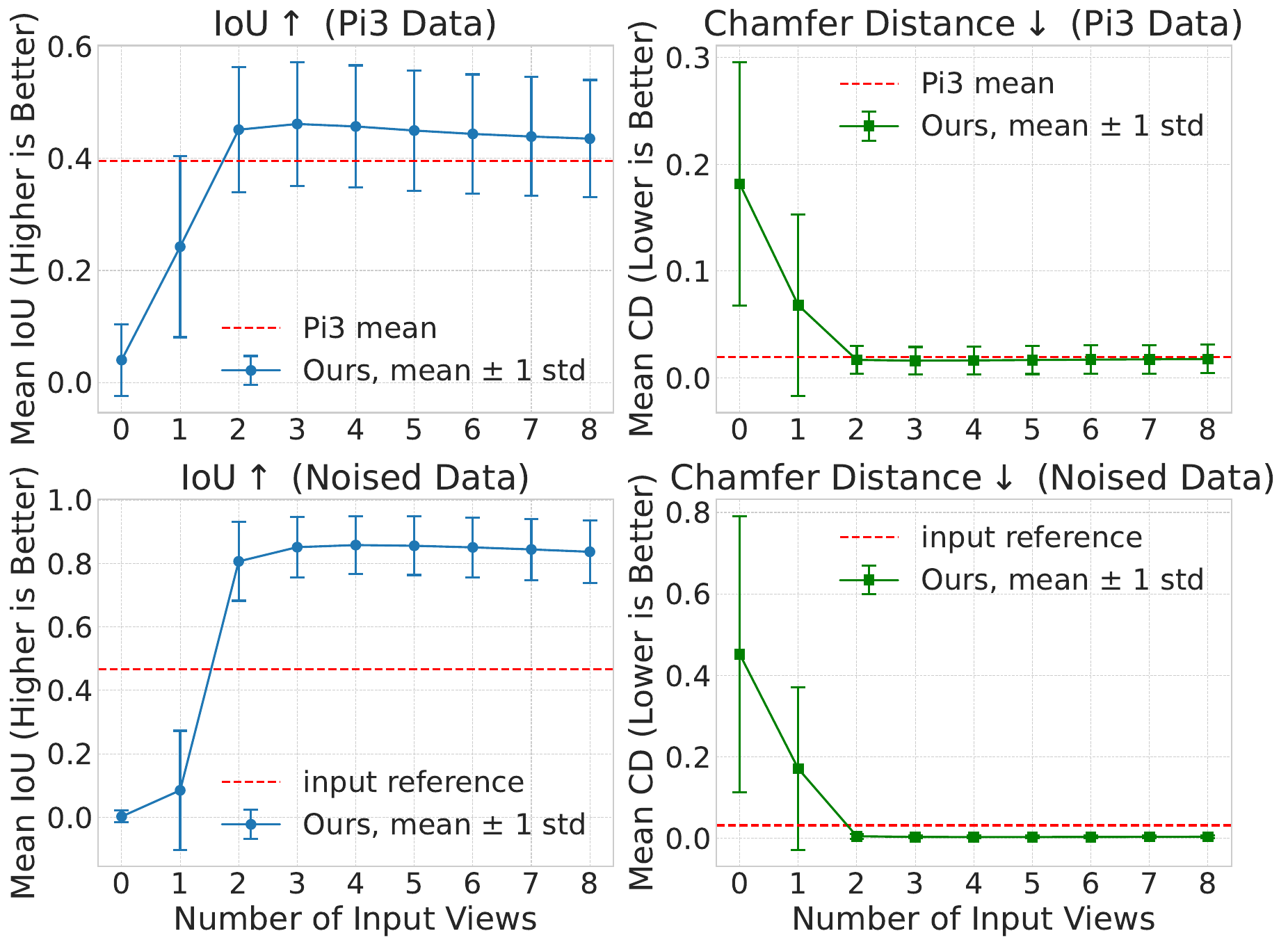}

   \caption{\textbf{Refinement Performance with Varying Input Views.} This plot shows the Mean IoU and Chamfer Distance on the Toys4k \cite{Toys4K21} dataset as the number of views varies from 0 to 8.}
   \label{fig:num_of_view}
\end{figure}

Our Hybrid Stream Transformer is designed to flexibly handle a variable number of input views. We analyze the impact of view count $S$, on reconstruction quality, with results on the Toys4k dataset \cite{Toys4K21} presented in Fig.~\ref{fig:num_of_view}. The plot reveals a non-monotonic relationship: performance improves significantly from a geometry-only baseline ($S=0$), peaks at 3 to 4 views, then slightly degrades.

We hypothesize this trend reflects a trade-off. Initially, adding views provides critical information for resolving geometric ambiguities. But an excess number of views leads to diminishing returns, while the accumulated noise from imperfect predictions may dilute the model's attention, making it harder to distill a single, coherent shape. 

\section{Application}
\label{sec:trellis_pipeline}

\begin{figure}[]
  \centering   \includegraphics[width=0.96\linewidth]{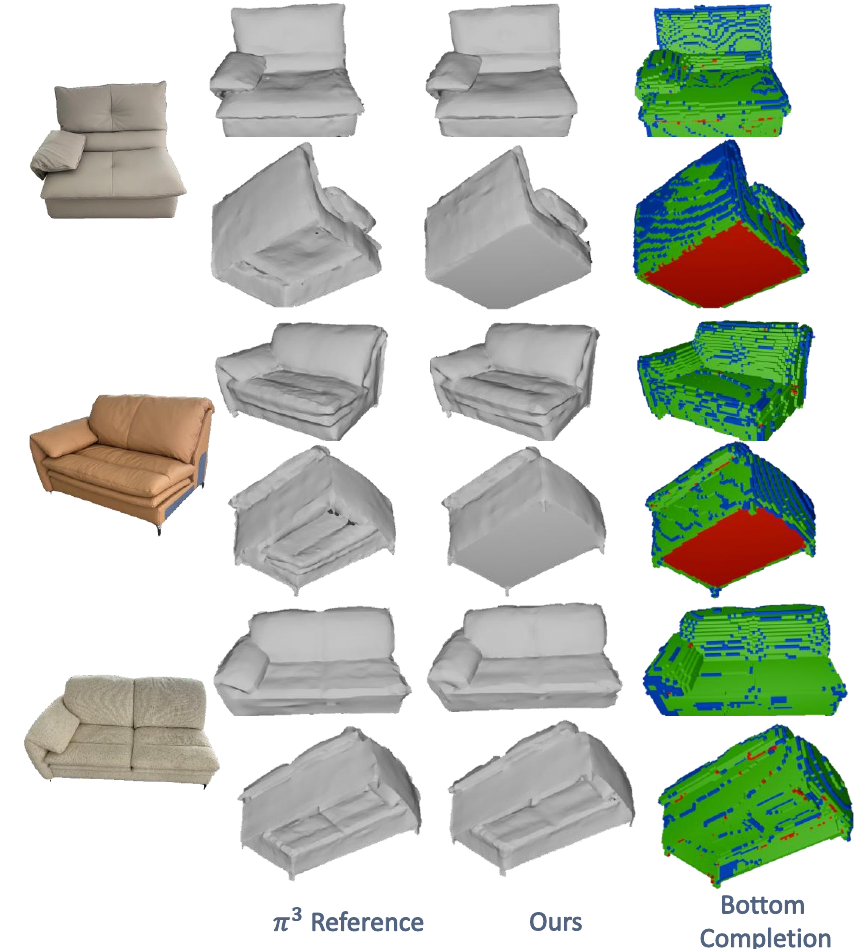}

   \caption{\textbf{Visualizing the Effect of \abbr's Refinement in Real-world 3D Sofa Creation Pipeline.} The ``Pi3 reference'' and ``Ours'' columns compare the final meshes produced by Trellis from the original and our refined voxels, respectively. The rightmost column displays the direct voxel modifications performed by our model (red: added, blue: removed).}
   \label{fig:sofa}
\end{figure}

We demonstrate \abbr's practical utility within a 3D asset creation pipeline using real-world sofa captures. In this setup, \abbr serves as a crucial bridge and is responsible for refining the coarse voxels generated by Pi3 \cite{pi3_25} from raw multi-view images.
Then the refined voxel grid is converted to the triangular mesh by the stage2 diffusion model of Trellis~\cite{trellis25}.
As depicted in Fig.~\ref{fig:sofa}, \abbr successfully reconstructs the unobserved undersides of the first two sofas while simultaneously refining the overall surface by eliminating fine-grained noise and pits. Consequently, when passed to Trellis, the resulting mesh exhibits a complete and smooth base.

However, the completion behavior of the model is not always consistent. For example, it does not reconstruct the base for the third sofa, treating it as an object that is inherently open from the bottom. We attribute this inconsistency to a fundamental ambiguity within our large-scale training data, where a significant portion of ground-truth objects also lack undersides. Furthermore, we observe the remaining holes in the final mesh, which arise from intricate gaps overlooked by \abbr, such as those beneath the sofa cushions. This suggests our current objective is not sufficiently sensitive to all local topological errors. Addressing this would likely require more explicit supervision, such as topology-aware loss functions or more geometrically complex training cases, which we leave as a promising direction for future work.
\section{Conclusion}
\label{sec:conclusion}

Our proposed \abbr is a framework for multi-view guided voxel refinement. Its effectiveness stems from a synergistic design: a Correctional Flow that leverages an imperfect but informative geometric prior, an Image Index to establish explicit spatial grounding, and a Hybrid Stream Transformer for robust feature fusion. Our experiments show that this integrated approach sets a new state-of-the-art in refining both realistic VFMs artifacts and severe synthetic corruptions.
Beyond these benchmarks, our application study (Sec.~\ref{sec:trellis_pipeline}) demonstrates the practical utility of VIAFormer as a crucial bridge in real-world 3D creation pipelines. We believe this work marks a pivotal step, paving the way for voxel-based methods to transition into the modern large-model, big-data paradigm.

{
    \small
    \bibliographystyle{ieeenat_fullname}
    \bibliography{main}
}

\clearpage
\setcounter{page}{1}
\maketitlesupplementary

\section{More Results}\label{sec:more_results}

\begin{figure*}[] 
    \centering 
    \includegraphics[width=1.0\textwidth]{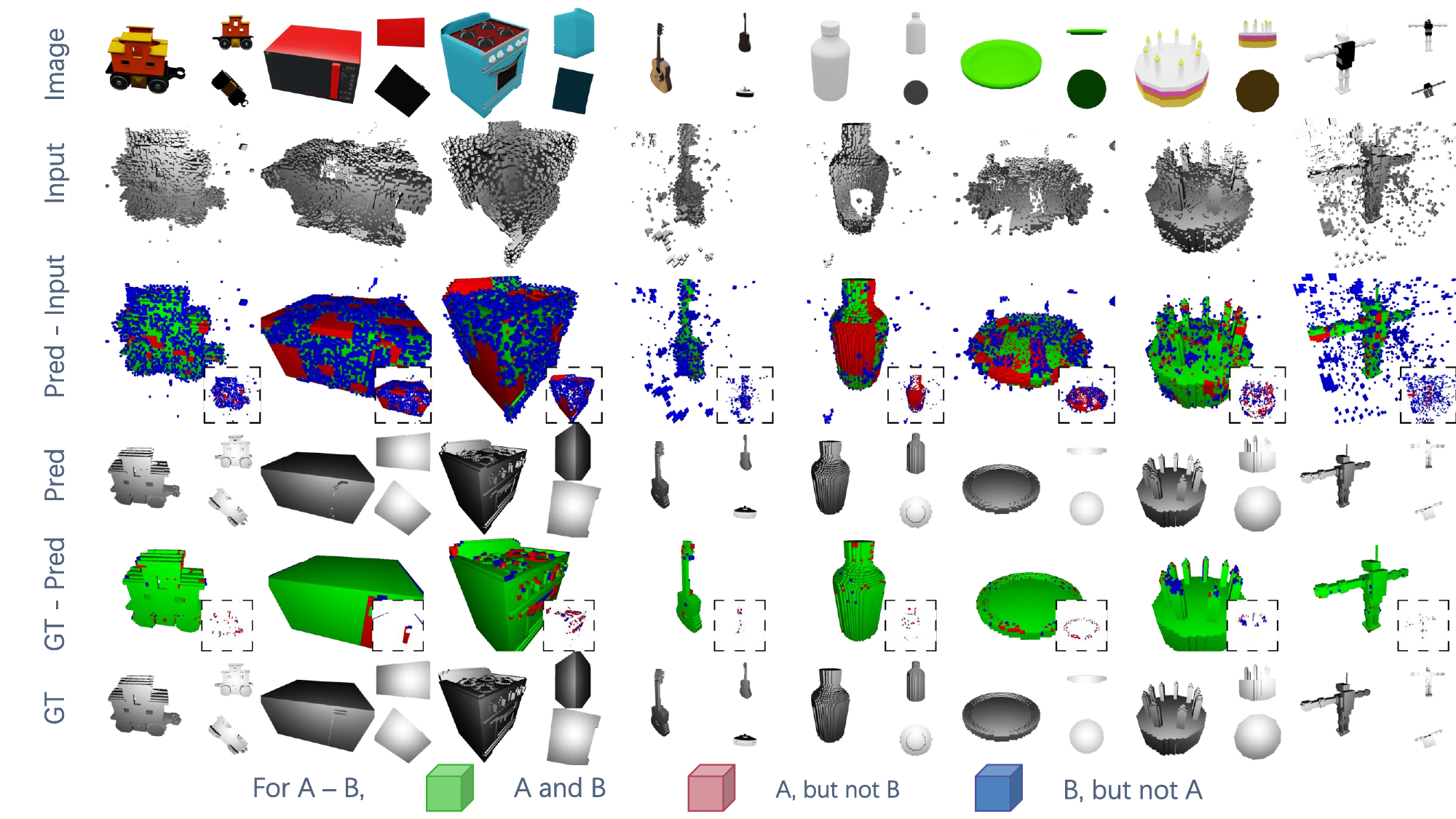} 
    \caption{\textbf{More refinement results on synthetic noise.} We present additional results demonstrating the \abbr's robustness across a diverse set of objects.}
    \label{fig:more_noised_results}
\end{figure*}

\begin{figure*}[] 
    \centering 
    \includegraphics[width=1.0\textwidth]{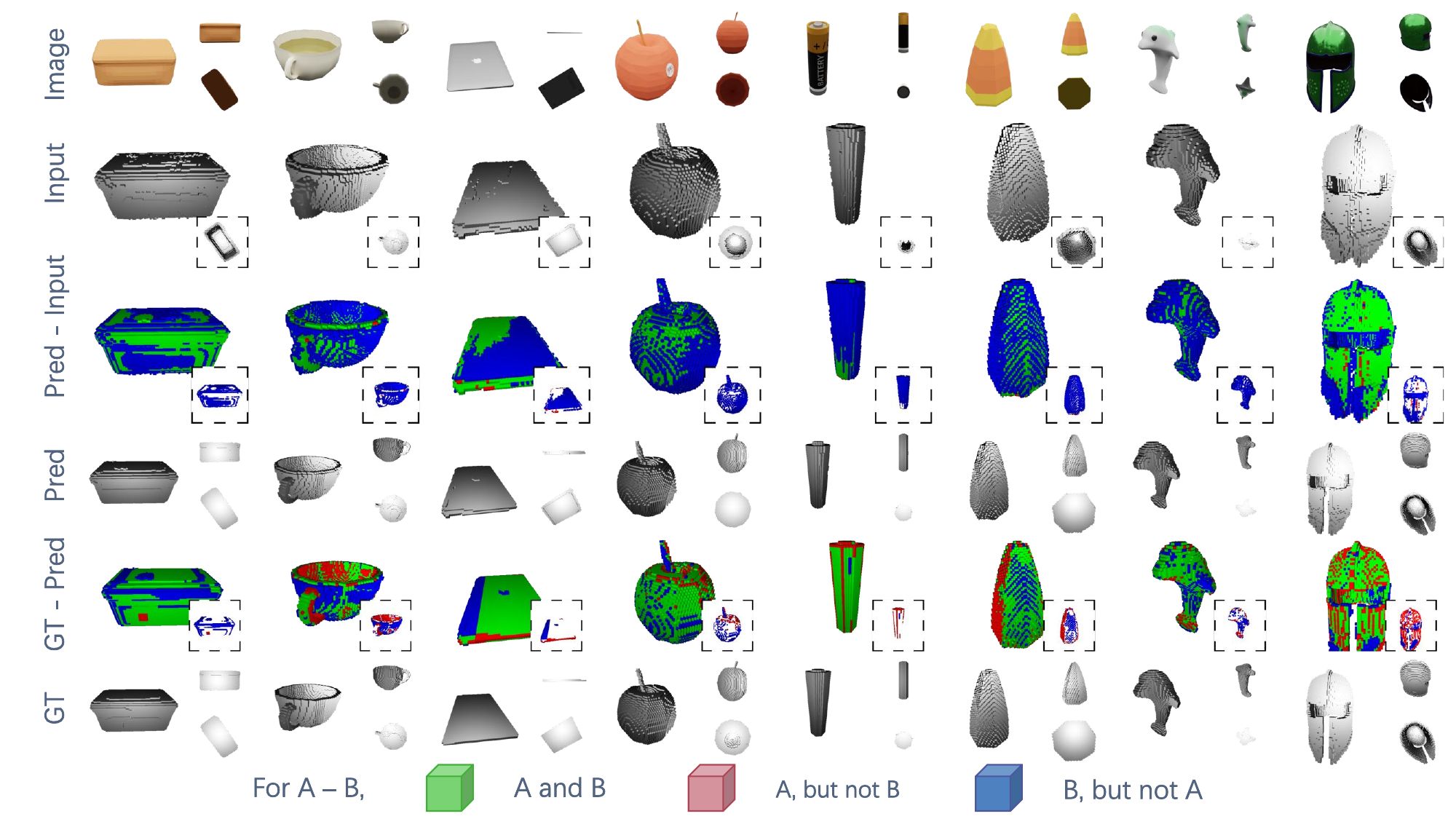} 
    \caption{\textbf{More refinement results on VFM-derived noise.} We present additional results demonstrating the \abbr's robustness across a diverse set of objects.}
    \label{fig:more_vfm_results}
\end{figure*}

To further demonstrate the robustness and versatility of our method, this section provides additional qualitative results. Fig.~\ref{fig:more_noised_results} and \ref{fig:more_vfm_results} showcase more refinement outcomes on synthetic and VFM-derived noise, respectively.

\begin{figure*}[] 
    \centering 
    \includegraphics[width=1.0\textwidth]{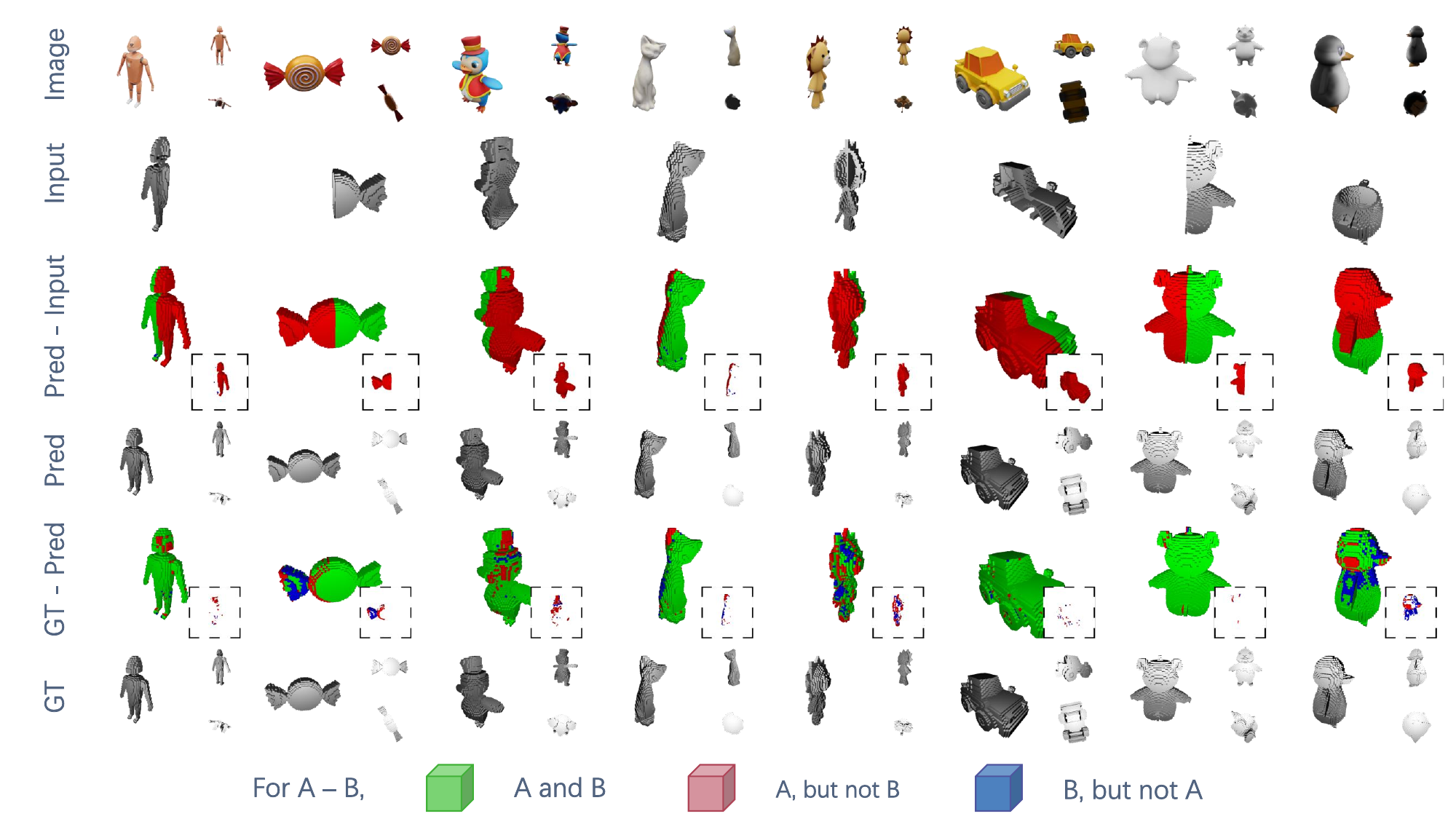} 
    \caption{\textbf{Qualitative results for the half-space completion task.} The model is tasked with completing a shape given half of the geometry (\textbf{Input}), 4 input views and VFM-estimated Image Index. The \textbf{Pred - Input} row visualizes this process, showing the provided half (green) and the generated half (red). The final predictions (\textbf{Pred}) and low residual errors (\textbf{GT - Pred}) demonstrate the model's ability to perform large-scale conditional generation by fusing geometric and visual information. The rendered \textbf{Image} is for reference only.} 
    \label{fig:halfspace_results}
\end{figure*}

Furthermore, we conduct an experiment to evaluate the model's capacity for half-space completion. The model is tasked with inferring the missing half of a shape, given the VFM-estimated image index ($\text{pos}_{I,i}$), four corresponding camera views ($\{I_i\}_{i=1}^4$), and one half of the ground-truth voxel grid ($v_{gt}$). The results, shown in Fig.~\ref{fig:halfspace_results}, indicate that our model is capable of reconstructing the missing geometry based on these inputs.

\section{Model Architecture Details}\label{sec:model_arch_detail}

\begin{figure*}[t] 
    \centering 
    \includegraphics[width=0.92\textwidth]{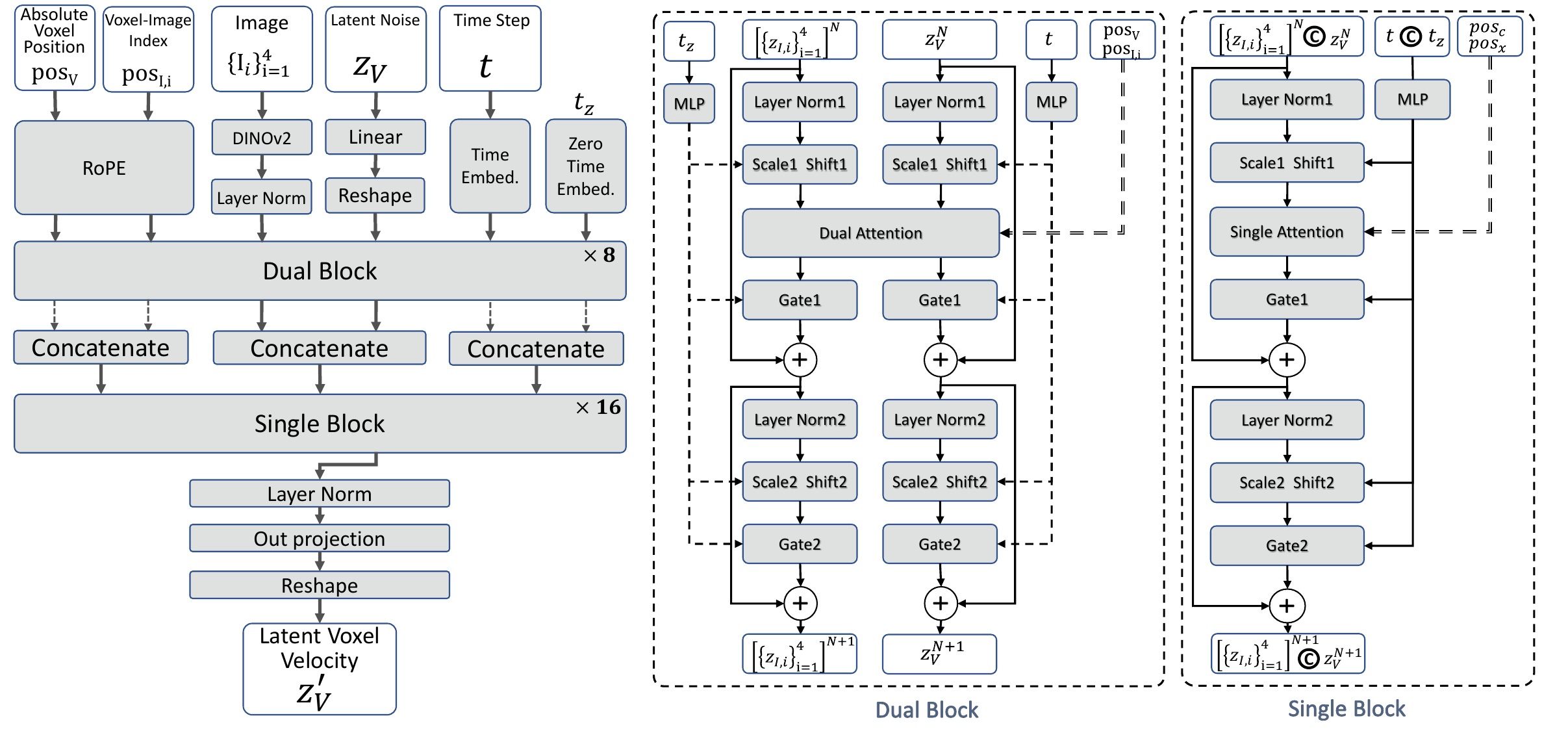} 
    \caption{\textbf{The \abbr Architecture.} Our model is a two-stage transformer. The first stage uses 8 Dual Blocks (middle) to process and align parallel voxel ($z_V$) and image ($\{z_{I,i}\}_{i=1}^S$) streams. The second stage concatenates these streams and uses 16 Single Blocks (right) for global refinement. The model outputs the final latent voxel velocity.}
    \label{fig:viaformer_arch}
\end{figure*}

\begin{figure*}[htbp]
    \centering 
    \begin{subfigure}[b]{0.39\textwidth}
        \includegraphics[width=\linewidth]{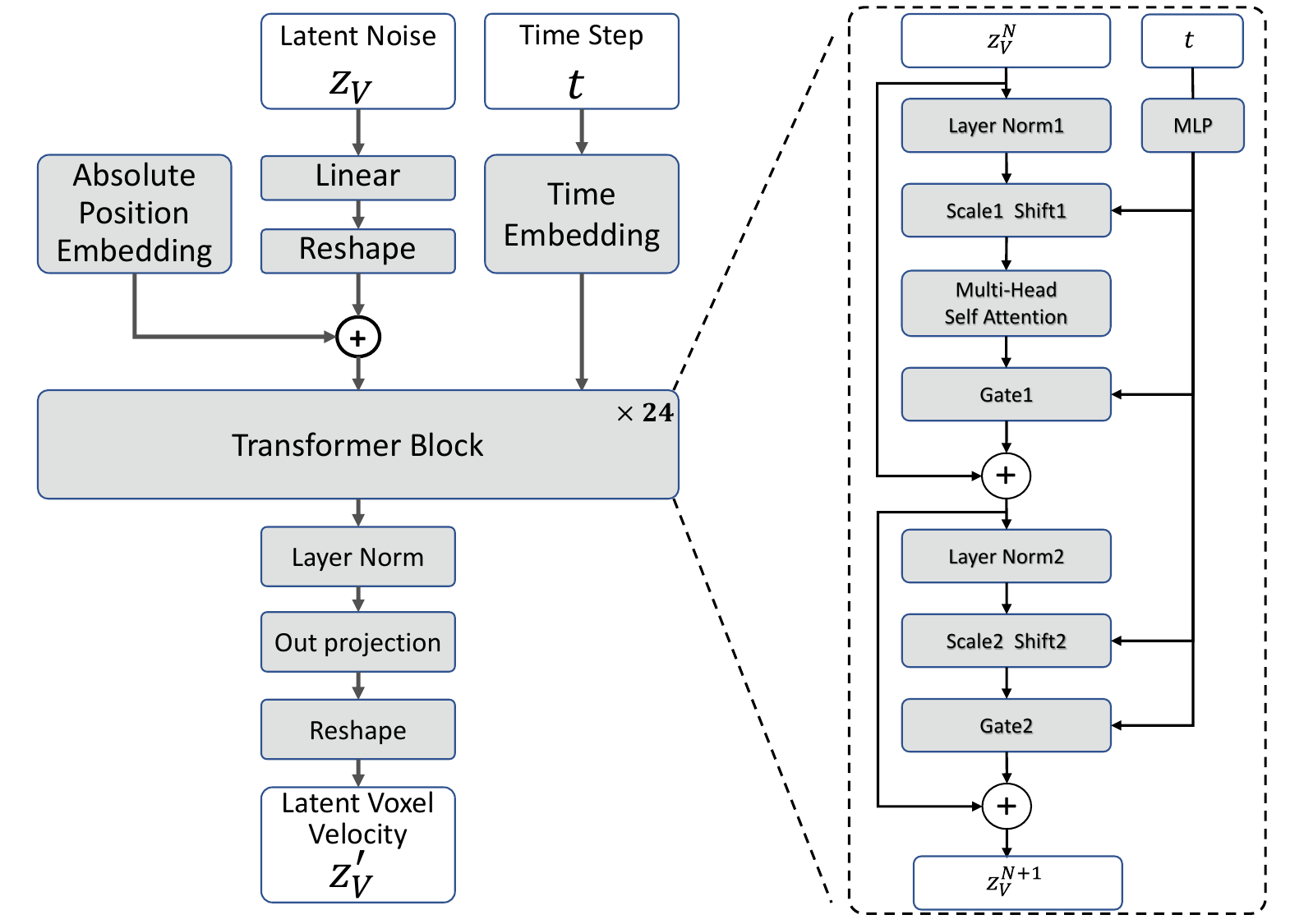}
        \caption{\textbf{The 24-Layer Self-Attention Baseline.} A geometry-only architecture that refines the noisy latent voxel grid ($z_V$) using a stack of 24 self-attention Transformer blocks, conditioned on time ($t$).}
        \label{fig:trellis_voxelonly_arch}
    \end{subfigure}%
    \hfill
    \begin{subfigure}[b]{0.59\textwidth}
        \includegraphics[width=\linewidth]{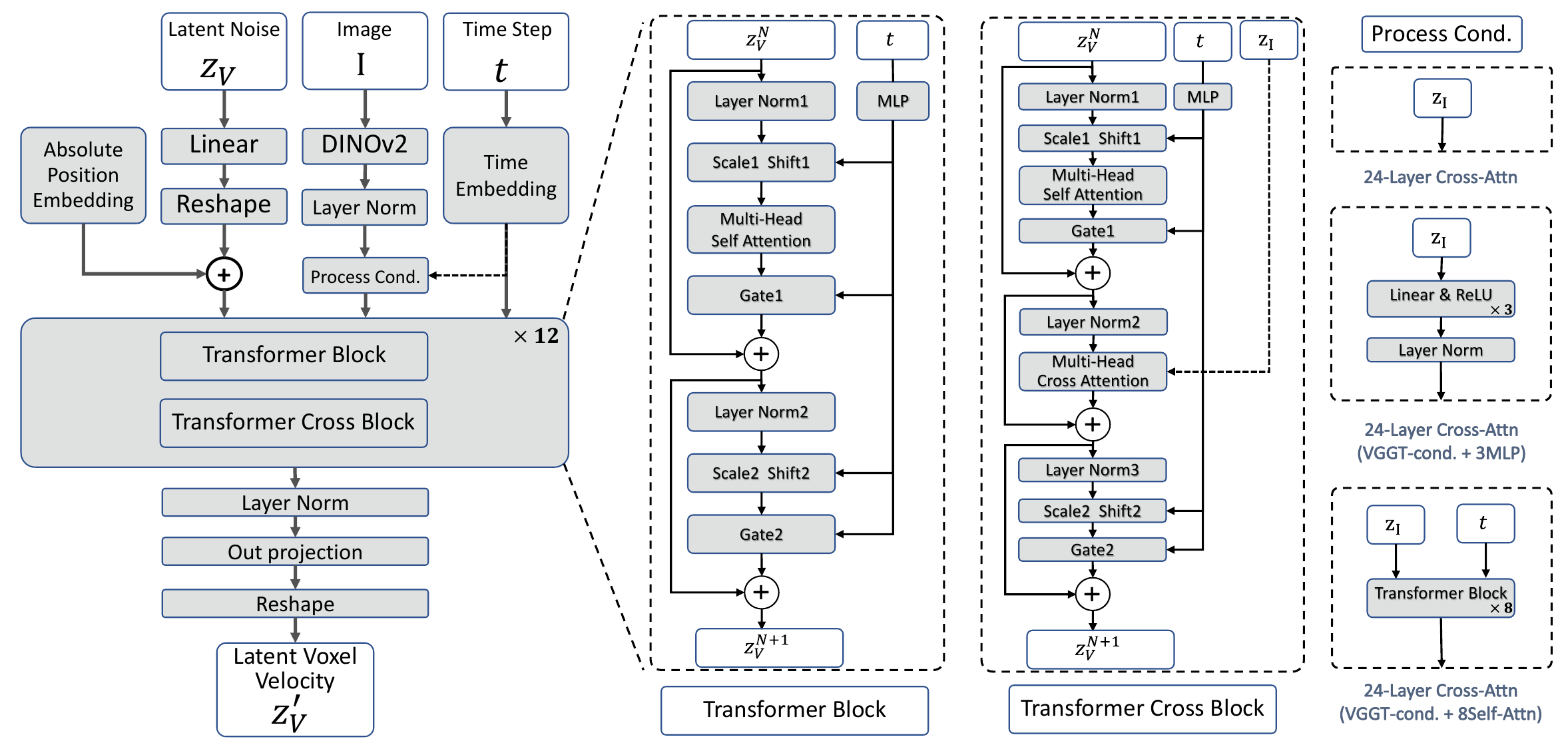}
        \caption{\textbf{The 24-Layer Cross-Attention Baseline.} An architecture composed of alternating self-attention and cross-attention blocks to fuse image conditioning ($z_I$) into the voxel stream ($z_V$). We test three conditioning variants (right panel): direct injection, a 3-layer MLP adapter, and an 8-layer self-attention adapter.}
        \label{fig:trellis_cross1view_arch}
    \end{subfigure}

    \caption{\textbf{Baseline architectures for ablation studies.}}
    \label{fig:trellis_model_arch}
\end{figure*}

This section provides detailed diagrams for the model architectures discussed in Sec.~\ref{sec:method}. Fig.~\ref{fig:viaformer_arch} illustrates the complete architecture of our proposed \abbr model. For the ablation studies, Fig.~\ref{fig:trellis_model_arch} details the baseline models used for comparison: (a) the geometry-only 24-Layer Self-Attention model, and (b) the 24-Layer Cross-Attention model with its various conditioning adapters.

\section{Data Process Details}

\begin{figure}[]
  \centering   \includegraphics[width=\linewidth]{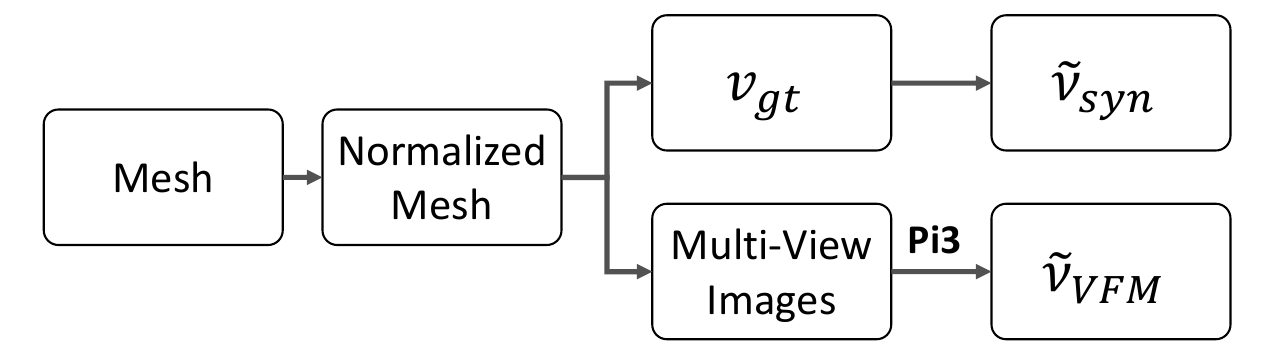}

   \caption{\textbf{Data process pipeline.} To simulate real-world noise and enhance model robustness, we generate two distinct types of voxel inputs: VFM Noise and Synthetic Corruptions.}
   \label{fig:data_process_pipeline}
\end{figure}

This section expands upon the data generation process from Sec.~\ref{sec:training_synthesis}, with a visual overview provided in Fig.~\ref{fig:data_process_pipeline}.

\subsection{VFM Noise}
We employ Pi3 \cite{pi3_25}, a powerful vision foundation model (VFM), to simulate point clouds that mimic real-world sensor data, which are typically coarse, noisy, and incomplete in occluded regions. For each sample in the dataset, we first exclude the bottom view from the set of rendered images, as such viewpoints are rarely feasible in practical capture scenarios.
The remaining multi-view images are then fed into Pi3 to predict per-view point clouds in the world coordinate system, along with confidence probabilities.
We apply a confidence threshold of $p>0.2$ to filter out low-confidence points, effectively removing background clutter.
The filtered point clouds from all views are fused to obtain a coarse reconstruction of the object’s geometry.
Since Pi3 produces outputs with arbitrary scale and orientation, we normalize the fused point cloud to unit space and align it using the ground-truth pose of the first view, thereby transforming the relative reconstruction into an absolute world coordinate frame.
Finally, the normalized point cloud is converted into a voxel grid at resolution $64$ using Open3D’s \texttt{create\_from\_point\_cloud\_within\_bounds} function. The resulting voxel representation is intentionally noisy and incomplete, forming a training pair with the clean ground-truth voxel grid, which is consistent with our voxel refinement objective.

\subsection{Synthetic Corruptions}

\begin{figure}[]
  \centering   \includegraphics[width=\linewidth]{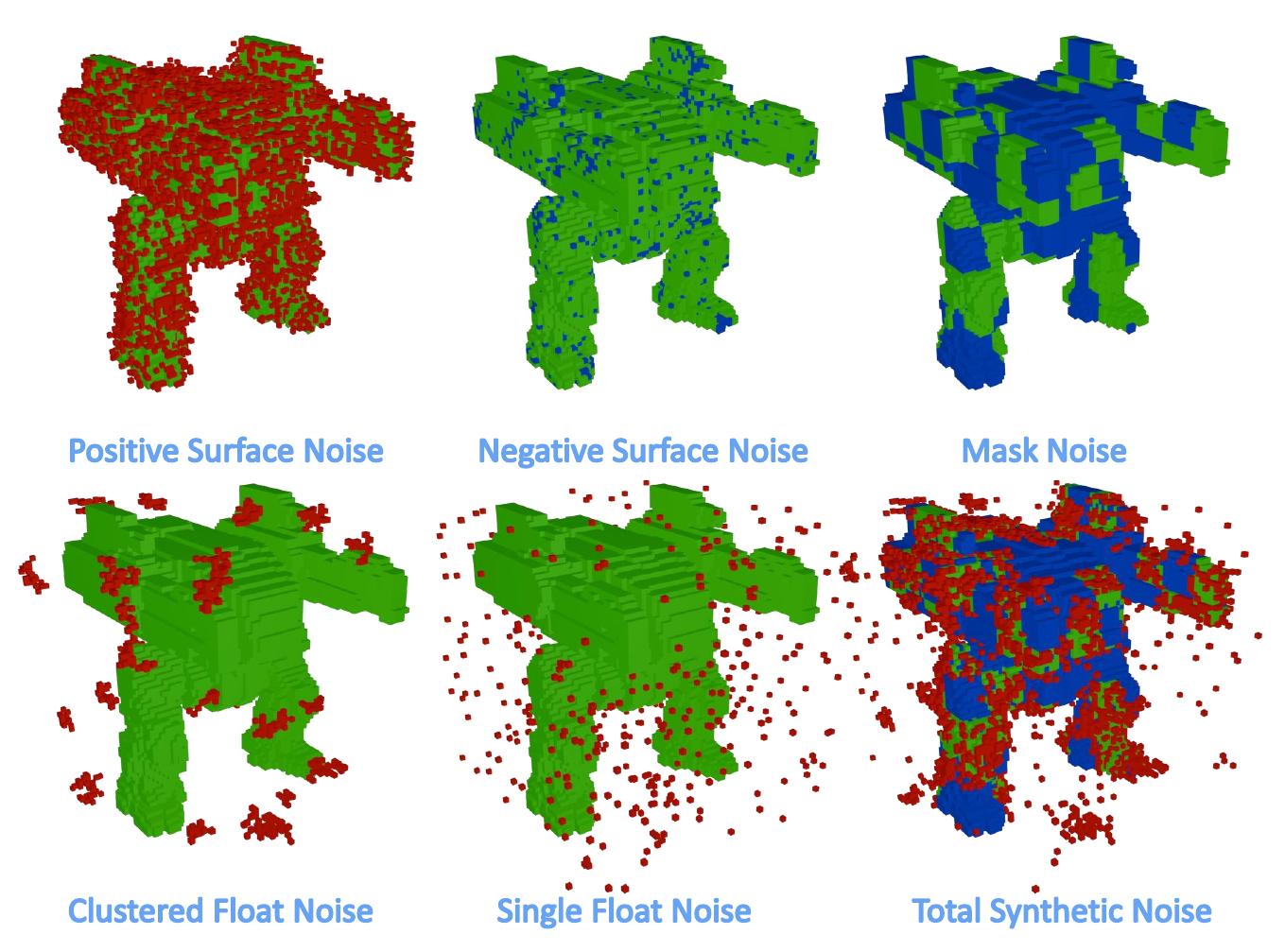}

   \caption{\textbf{Synthetic noise generation modules.} A clean model (green) is corrupted with various noise types (additions in red, removals in blue). Top row: Positive Surface, Negative Surface, and Mask Noise. Bottom row: Clustered Float, Single Float, and the combined Total Synthetic Noise used for training.}
   \label{fig:synthesic_noise}
\end{figure}

To complement the realistic VFM noise and further improve model robustness, we introduce a diverse synthetic corruption pipeline. We argue that existing approaches like block-wise masking used in PatchComplete \cite{PatchComplete22} are too limited for general-purpose refinement. Therefore, our pipeline explicitly adds surface noise and volumetric artifacts (floaters) to better prepare the model for a wider range of real-world degradations.

As shown in Fig.~\ref{fig:synthesic_noise}, we design our synthetic corruptions as composable noise modules. Each module takes an occupancy grid $v \in \{0,1\}^{R \times R \times R}$ as input, along with specific control parameters, and generates an additive noise score map $n \in \mathbb{R}^{R \times R \times R}$. The noise scores from all applied modules are summed and added to the original grid, which is then binarized to produce the final corrupted grid $\tilde{v}$:
\begin{equation}
\tilde{v} = \text{clamp}\left(v + \sum n, 0, 1\right).
\end{equation}

\noindent\textbf{1) SDF-based Single-Voxel Noise.}
This module generates isolated noise voxels based on their distance to the object's surface, calculated via a Signed Distance Function (SDF). Voxels within a specified SDF range (a ``shell'') become candidates. Each candidate is then selected with a probability $p$, randomly sampled from a predefined list (see Table~\ref{tab:noise_params}). This process is used to create both:
\begin{itemize}
    \item \textbf{Positive Noise} (e.g., \textit{Surface Noise}, \textit{Far-field Noise}), which adds spurious voxels outside the object surface.
    \item \textbf{Negative Noise} (e.g., \textit{Surface Erosion}), which creates holes by removing voxels just inside the surface.
\end{itemize}

\noindent\textbf{2) SDF-based Clustered Noise.}
To simulate more spatially coherent artifacts, this module extends the SDF-based approach. Instead of adding or removing single voxels, it treats each stochastically selected point as a seed. From each seed, a cluster of $k$ connected voxels is grown, where $k$ is randomly sampled from a list. This generates larger, contiguous artifacts such as \textit{Clustered Floaters} (detached fragments) or \textit{Clustered Erosion} (larger cavities).

\noindent\textbf{3) Coarse-level Masking.}
This module simulates large-scale, structural incompleteness by removing entire blocks of voxels. It achieves this by generating a low-resolution binary mask, upscaling it to the full grid resolution, and applying a large negative score to remove the voxels in the masked-out regions.

\noindent\textbf{4) Half-Space Removal.}
Distinct from our main noise pipeline, Half-Space Removal is a special corruption applied exclusively to a separate training subset. This drastic method, which deterministically removes all voxels on one side of a plane, is designed to rigorously test the model's large-scale, visually-conditioned generative capabilities. These samples are not mixed with those receiving VFM or other synthetic noises.

\begin{table*}[t]
\centering
\begin{threeparttable}

\caption{\textbf{Hyperparameters for Synthetic Corruption Modules.} This table details the parameters for the noise modules used in our training pipeline. For each corruption pass, a specific value is randomly sampled from the corresponding parameter list. The base resolution is $64^3$.}
\label{tab:noise_params}

\begin{tabular}{@{}llccc@{}}
\toprule
\textbf{Noise Type} & \textbf{Description} & \textbf{SDF Range} & \textbf{Probability List ($p$)} & \textbf{Cluster Size List ($k$)} \\ \midrule
\multicolumn{5}{l}{\textit{\textbf{Positive Noise (Floaters)}}} \\
\quad Surface Noise & Single voxel addition & $(0, 0.04]$ & \tnote{a} & --- \\
\quad Surface Noise & Single voxel addition & $(0.04, 0.08]$ & \tnote{b} & --- \\
\quad Far-field Noise & Single voxel addition & $(0.15, 2.0]$ & \tnote{c} & --- \\
\quad Clustered Floaters & Multi-voxel addition & $(0.15, 2.0]$ & \tnote{d} & \tnote{f} \\ \midrule
\multicolumn{5}{l}{\textit{\textbf{Negative Noise (Holes)}}} \\
\quad Surface Erosion & Single voxel removal & $(0, 0.04]$ & \tnote{a} & --- \\
\quad Surface Erosion & Single voxel removal & $(0.04, 0.08]$ & \tnote{b} & --- \\
\quad Clustered Erosion & Multi-voxel removal & $(0, 0.04]$ & \tnote{e} & \tnote{f} \\ \midrule
\multicolumn{5}{l}{\textit{\textbf{Masking Noise}}} \\
\quad Coarse Masking & \begin{tabular}[c]{@{}l@{}}Block removal at\\ various resolutions\end{tabular} & --- & \begin{tabular}[c]{@{}c@{}}Foreground prob.\\ in $[0.5, 1.0]$\end{tabular} & \begin{tabular}[c]{@{}c@{}}Mask Res.\\ $\{16, 8, 8, 4\}$\end{tabular} \\ \bottomrule
\end{tabular}

\begin{tablenotes}
    \item[\textbf{Parameter Lists:}]
    \item[a] $\{0.01, 0.02, 0.03, 0.04, 0.05, 0.07, 0.10, 0.15, 0.20, 0.25, 0.30, 0.35, 0.40\}$
    \item[b] $\{0.005, 0.01, 0.02, 0.03, 0.04, 0.05, 0.08, 0.12, 0.15, 0.20\}$
    \item[c] $\{10^{-4}, 2.5\times 10^{-4}, 5\times 10^{-4}, 10^{-3}, 2\times 10^{-3}\}$
    \item[d] $\{10^{-5}, 2\times 10^{-5}, 5\times 10^{-5}, 10^{-4}\}$
    \item[e] $\{8\times 10^{-4}, 2\times 10^{-3}, 5\times 10^{-3}, 8\times 10^{-3}\}$
    \item[f] $\{8, 10, 12, 14, 16, 18, 20, 25, 30\}$
\end{tablenotes}

\end{threeparttable}
\end{table*}

\subsection{Additional Preprocessing in Real-World Scenarios}
Compared to synthetic rendering datasets, real-world applications often lack ground-truth information, resulting in incomplete inputs that require additional preprocessing. To address this challenge, we leverage some large powerful model to estimate missing information. Specifically, we employ Pi3~\cite{pi3_25} to predict camera parameters from real captured images, SAM2~\cite{ravi2024sam2} to generate precise object segmentation masks, OrientAnything~\cite{orient_anything} to estimate object orientation for canonical pose normalization, and StableNormal~\cite{ye2024stablenormal} to infer surface normal maps from RGB inputs.

\subsection{K-Means View Selection Algorithm}
\label{appendix:kmeans_view_selection}
Algo.~\ref{alg:view_selection} details the K-Means view selection strategy mentioned in Sec.~\ref{sec:experimental_setup}. The objective is to select a diverse set of $S$ views by clustering their angular positions. This method assumes all cameras point towards the origin, enabling the direct use of their 3D positions to derive angular features (longitude and latitude) for clustering. The algorithm partitions candidate views into $S$ clusters and selects the most central view from each to ensure maximal angular coverage. For our experiments, total view count $N=150$, the pitch angle for candidate views is constrained between $\theta_{\text{min}} = -10^\circ$ and $\theta_{\text{max}} = 30^\circ$.

\begin{algorithm}[h!]
\caption{K-Means View Selection}
\label{alg:view_selection}
\begin{algorithmic}[1]
\Require 
    A set of $N$ camera positions $\mathcal{P} = \{p_i\}_{i=1}^N$, where $p_i = (x_i, y_i, z_i) \in \mathbb{R}^3$.
\Require
    The desired number of views $S$.
\Require
    Minimum and maximum pitch angles, $\theta_{\text{min}}$ and $\theta_{\text{max}}$.
\Ensure 
    A set of $S$ selected view indices $\mathcal{I}_{\text{selected}}$.
\Statex \textbf{Assumption:} All cameras are oriented to look at the origin $(0,0,0)$.

\State Initialize an empty list for angular features $\mathcal{F} \leftarrow []$.
\For{each camera position $p_i = (x_i, y_i, z_i)$ in $\mathcal{P}$}
    \State $\text{lng}_i \leftarrow \operatorname{atan2}(y_i, x_i)$ \Comment{Calculate azimuth}
    \State $\text{lat}_i \leftarrow \operatorname{atan2}(z_i, \sqrt{x_i^2 + y_i^2})$ \Comment{Calculate pitch}
    \State Append $(\text{lng}_i, \text{lat}_i)$ to $\mathcal{F}$.
\EndFor

\State $\mathcal{I}_{\text{cand}} \leftarrow \{i \mid \theta_{\text{min}} \le \mathcal{F}[i]_{\text{lat}} \le \theta_{\text{max}}\}$ \Comment{Filter views}
\If{$|\mathcal{I}_{\text{cand}}| < S$}
    \State $\mathcal{I}_{\text{cand}} \leftarrow \{1, 2, \dots, N\}$ \Comment{Fallback if too few views}
\EndIf

\State $\mathcal{F} \leftarrow \{\mathcal{F}[i] \mid i \in \mathcal{I}_{\text{cand}}\}$

\State Generate a random angle $\phi_{\text{rand}} \sim U(0, 2\pi)$.
\For{each index $i \in \{1, \dots, N\}$}
    \State $\mathcal{F}[i]_{\text{lng}} \leftarrow (\mathcal{F}[i]_{\text{lng}} + \phi_{\text{rand}}) \pmod{2\pi}$
\EndFor

\State $(\mathcal{C}_1, \dots, \mathcal{C}_S), (\mu_1, \dots, \mu_S) \leftarrow \operatorname{KMeans}(\mathcal{F}, S)$ 
\Statex \Comment{$\mathcal{C}_k$ are clusters of feature vectors, $\mu_k$ are centroids}

\State $\mathcal{I}_{\text{selected}} \leftarrow \emptyset$.
\For{$k=1$ to $S$}
    \State Find original index $j^*$ of the feature in $\mathcal{F}$ that is closest to centroid $\mu_k$:
    \Statex \qquad $j^* \leftarrow \underset{j \in \mathcal{I} \text{ s.t. the view with index } j \text{ is in cluster } \mathcal{C}_k}{\operatorname{argmin}} \| \mathcal{F}[j] - \mu_k \|_2$
    \State Add $j^*$ to $\mathcal{I}_{\text{selected}}$.
\EndFor

\State \Return $\mathcal{I}_{\text{selected}}$
\end{algorithmic}
\end{algorithm}

\section{Evaluation Metrics}

\noindent\textbf{1) IoU (Intersection over Union):}  IoU measures the overlap between the predicted voxel set $\hat{v}$ and the ground-truth set $v_{\text{gt}}$:
\begin{equation}
\label{eq:iou}
\text{IoU}(\hat{v}, v_{\text{gt}}) = \frac{|\hat{v} \cap v_{\text{gt}}|}{|\hat{v} \cup v_{\text{gt}}|}.
\end{equation}

\noindent\textbf{2) Chamfer Distance (CD):} To evaluate surface-level fidelity, we compute the Chamfer Distance. First, the predicted and ground-truth voxel grids are normalized to fit within a unit cube of $[-0.5, 0.5]^3$. We then extract their point clouds of grid $\mathrm{P}(\hat{v})$ and $\mathrm{P}(v_{\text{gt}})$. The CD is defined as the mean squared closest point distance between these two sets:
\begin{equation}
\label{eq:cd}
\begin{split}
    \mathrm{CD}(\hat{v}, v_{\text{gt}}) = & \frac{1}{|\hat{v}|} \sum_{x \in \mathrm{P}(\hat{v})} \min_{y \in \mathrm{P}(v_{\text{gt}})} \|x - y\|_2^2 \\
    & + \frac{1}{|v_{\text{gt}}|} \sum_{y \in \mathrm{P}(v_{\text{gt}})} \min_{x \in \mathrm{P}(\hat{v})} \|y - x\|_2^2.
\end{split}
\end{equation}

\section{Attention Map} \label{appendix:attention_map}

\subsection{Attention Map Theory} \label{appendix:attention_map_theory}

This section provides a theoretical overview of the Transformer attention mechanism, focusing on the generation and interpretation of attention maps. This foundation is essential for understanding the attention map visualizations presented in  Fig.~\ref{fig:attention_map} and our diagnosis of ``Attention Collapse'' in Sec.~\ref{sec:necessity_of_union_space}.

In a Transformer attention layer, input tokens are first projected into three distinct representations: Query ($Q$), Key ($K$), and Value ($V$). For self-attention, these projections originate from the same input sequence. For cross-attention, $Q$ is projected from the target sequence (e.g., voxels), while $K$ and $V$ are projected from the source sequence (e.g., image patches). Let the target sequence have length $L_q$ and the source sequence have length $L_{kv}$. The input tensors are typically of shapes $\mathbb{R}^{B \times L_q \times C}$ and $\mathbb{R}^{B \times L_{kv} \times C}$ respectively.

For multi-head attention, the feature dimension $C$ is divided among $H$ parallel heads, with each head having a dimension of $d_k = C/H$. The $Q, K, V$ tensors are then reshaped and transposed to isolate the heads for computation, resulting in the shapes:
\begin{equation}
    Q \in \mathbb{R}^{B \times H \times L_q \times d_k}, \quad K, V \in \mathbb{R}^{B \times H \times L_{kv} \times d_k}.
\end{equation}

The core of the mechanism is the scaled dot-product attention formula, which computes the output as a weighted sum of $V$:
\begin{equation}
\text{Attention}(Q, K, V) = \text{softmax}\left(\frac{QK^T}{\sqrt{d_k}}\right)V.
\end{equation}
The component used for visualization is the attention map, which is the tensor of weights that pre-multiplies $V$:
\begin{equation}
\text{AttentionMap} = \text{softmax}\left(\frac{QK^T}{\sqrt{d_k}}\right)\in\mathbb{R}^{B\times H\times L_q\times L_{kv}}.
\end{equation}
Its role is to weight $V$ by multiplication of the batch matrix to produce the final attention output:
\begin{equation}
\underbrace{\text{AttentionMap}}_{[B, H, L_q, \mathbf{L_{kv}}]} \ @ \ \underbrace{V}_{[B, H, \mathbf{L_{kv}}, d_k]} \ \Rightarrow \ \underbrace{\text{AttentionOutput}}_{[B, H, L_{q}, d_k]}.
\end{equation}
This structure clarifies the role of the attention map: its element at $(i,j)$ represents the weight assigned by the $i$-th query token to the $j$-th key token, thereby encoding a learned, directional relationship across modalities.

In visually intuitive domains such as image processing, it is common practice to visualize attention from a specific query token. The corresponding row of the attention map is extracted and reshaped back into a heatmap to illustrate that the model's attention is focused on semantically meaningful patches. However, \abbr  operates within an abstract latent space—the velocity field defined by the Correctional Flow. This space lacks a direct, visually interpretable mapping. Consequently, we present the full, abstract $L_q\times L_{kv}$ attention maps directly. The goal of our visualizations is not to pinpoint focus on specific spatial locations, but rather to diagnose the global patterns of the attention mechanism, which is critical for identifying \textbf{Attention Collapse} case discussed in our analysis.

\subsection{Attention Collapse Evidence}

\begin{figure}[]
  \centering   \includegraphics[width=\linewidth]{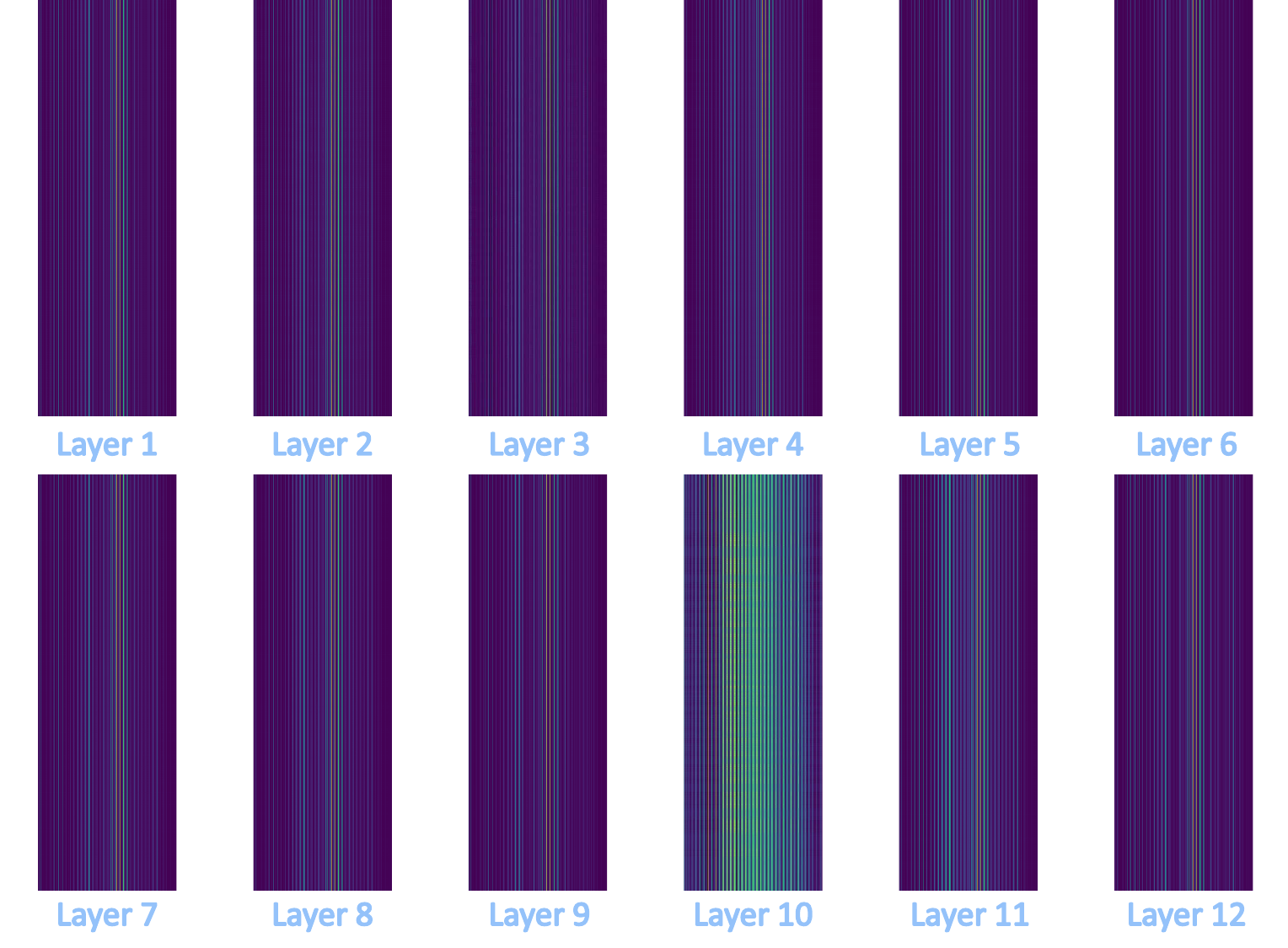}

   \caption{\textbf{Full visualization of Attention Collapse in the cross-attention baseline.} This figure displays the mean attention map (averaged across 16 heads) for each of the 12 cross-attention layers.}
   \label{fig:attention_collapse_full}
\end{figure}

This section provides the full visual evidence for the ``Attention Collapse'' claim (Sec.~\ref{sec:necessity_of_union_space}) in Fig.~\ref{fig:attention_collapse_full}, which shows the consistent uniform vertical stripe pattern in the mean attention maps for all 12 cross-attention layers of our baseline, a pattern indicating that every voxel token attends non-selectively to all image tokens, confirming the systemic failure to learn spatial correspondence.

\end{document}